\documentclass[journal]{IEEEtran}

\ifCLASSOPTIONcompsoc
  \usepackage[nocompress]{cite}
\else
  \usepackage{cite}
\fi
\ifCLASSINFOpdf
\else
\fi

\pdfoutput=1

\usepackage{subfigure}
\usepackage{xcolor}
\usepackage{footnote}
\usepackage{multirow}
\usepackage{times}
\usepackage{epsfig}
\usepackage{graphicx}
\usepackage{amsmath}
\usepackage{amssymb}
\usepackage{fixltx2e}
\usepackage{color}
\usepackage{algorithm}
\usepackage{algpseudocode}
\usepackage{hyperref}
\usepackage[switch]{lineno} 

\usepackage{makecell}
\usepackage{footnote}
\usepackage{subfigure}
\usepackage{multicol}
\newcommand\red[1]{{\color{black}#1}}

\graphicspath{{./figures/}{./figures/draw/}}

\begin{document}
%\linenumbers
\title{Simultaneous Feature Aggregating and Hashing for Compact Binary Code Learning}

%%%%%%%%%%%%%%%%%%%%%%%%%%%%%%%%%%%%%%%%%%%%
%%% uncomment for showing author list
\author{%Thanh-Toan Do, Khoa Le, Tuan Hoang, Huu Le, Tam Nguyen, Ngai-Man Cheung, Ian Reid
Thanh-Toan Do, Khoa Le, Tuan Hoang, Huu Le, Tam V. Nguyen, Ngai-Man Cheung
\IEEEcompsocitemizethanks{
\IEEEcompsocthanksitem 
Thanh-Toan Do is with the University of Liverpool, United Kingdom.
%\protect\\
%E-mail: \{thanh-toan.do, ian.reid\}@adelaide.edu.au
E-mail: thanh-toan.do@liverpool.ac.uk
\IEEEcompsocthanksitem Khoa Le, Tuan Hoang, and Ngai-Man Cheung are with the Singapore University of Technology and Design, Singapore.
%\protect\\
E-mail: \{letandang\_khoa,ngaiman\_cheung\}@sutd.edu.sg, nguyenanhtuan\_hoang@mymail.sutd.edu.sg
\IEEEcompsocthanksitem Huu Le is with Chalmers University of Technology, Sweden.
%\protect\\
E-mail: huulem@outlook.com
\IEEEcompsocthanksitem Tam V. Nguyen is with the University of Dayton, United States.
E-mail: \protect\\{tamnguyen}@udayton.edu
}
}

%%%%%%%%%%%%%%%%%%%%%%%%%%%%%%%%%%%%%%%%%%%%%%%%%%

%{Shell \MakeLowercase{\textit{et al.}}: Bare Demo of IEEEtran.cls for Computer Society Journals}

\IEEEtitleabstractindextext{%

%\begin{IEEEkeywords}
%Computer Society, IEEE, IEEEtran, journal, \LaTeX, paper, template.
%\end{IEEEkeywords}
}

\maketitle

\IEEEdisplaynontitleabstractindextext

\IEEEpeerreviewmaketitle
\renewcommand{\algorithmicrequire}{\textbf{Input:}}
\renewcommand{\algorithmicensure}{\textbf{Output:}}
\renewcommand{\algorithmicrequire}{\textbf{Input:}}
\renewcommand{\algorithmicensure}{\textbf{Output:}}

\def\x{{\mathbf x}}
\def\y{{\mathbf y}}
\def\w{{\mathbf w}}
\def\h{{\mathbf h}}
\def\z{{\mathbf z}}
\def\cc{{\mathbf c}}
\def\bb{{\mathbf b}}
\def\I{{\mathbf 1}}
\def\q{{\mathbf q}}
\def\h{{\mathbf h}}
\def\f{{\mathbf f}}

\def\B{{\mathbf B}}
\def\Y{{\mathbf Y}}
\def\Ro{{\mathbf R}}
\def\X{{\mathbf X}}
\def\Y{{\mathbf Y}}
\def\R{{\mathbb R}}
\def\H{{\mathbf H}}
\def\Z{{\mathbf Z}}
\def\W{{\mathbf W}}
\def\V{{\mathbf V}}
\def\Q{{\mathbf Q}}
\def\I{{\mathbf I}}
\def\1{{\mathbf 1}}
\def\S{{\mathbf S}}
\def\V{{\mathbf V}}
\def\P{{\mathbf P}}
\def\Vs{{\mathcal V}}
\def\O{{\mathcal O}}
\def\0{{\mathbf 0}}

\newcommand\norm[1]{\left\lVert#1\right\rVert}
\newcommand{\overbar}[1]{\mkern 1mu\overline{\mkern-1mu#1\mkern-1mu}\mkern 1mu}

\newtheorem{theorem}{Theorem}[section]
\newtheorem{lemma}[theorem]{Lemma}
\newtheorem{proposition}[theorem]{Proposition}
\newtheorem{corollary}[theorem]{Corollary}
\newtheorem{definition}[theorem]{Definition}

\newenvironment{proof}[1][Proof]{\begin{trivlist}
\item[\hskip \labelsep {\bfseries #1}]}{\end{trivlist}}
\newenvironment{example}[1][Example]{\begin{trivlist}
\item[\hskip \labelsep {\bfseries #1}]}{\end{trivlist}}
\newenvironment{remark}[1][Remark]{\begin{trivlist}
\item[\hskip \labelsep {\bfseries #1}]}{\end{trivlist}}

\begin{abstract}
 
Representing images by compact hash codes is an attractive approach for  large-scale content-based image retrieval.  In most state-of-the-art hashing-based image retrieval systems, for each image, local descriptors are first aggregated as a global representation vector.
This global vector is then subjected to a hashing function to generate a binary hash code. 
In previous works, the aggregating and the hashing processes are designed independently. 
Hence these frameworks may generate suboptimal hash codes.
In this paper, we first propose a novel unsupervised hashing framework in which feature aggregating and hashing are designed simultaneously and optimized jointly. 
Specifically, our joint optimization generates aggregated representations that can be better reconstructed by some binary codes.  This leads to more discriminative binary hash codes and improved retrieval accuracy. In addition, the proposed method is flexible. It can be extended for supervised hashing. When the data label is available, the framework can be adapted to learn binary codes which minimize the reconstruction loss w.r.t. label vectors. 
Furthermore, we also propose a fast version of the state-of-the-art hashing method
Binary Autoencoder to be used in our proposed frameworks. Extensive experiments on benchmark datasets under various settings show that the proposed methods outperform  state-of-the-art unsupervised and supervised hashing methods.

\end{abstract}
\begin{IEEEkeywords}
Image search, binary hashing, aggregating, embedding.
\end{IEEEkeywords}

\section{Introduction}
 
Content-based image retrieval is an important problem in computer vision with many applications, including visual search~\cite{DBLP:conf/cvpr/JegouZ14,DBLP:conf/cvpr/ArandjelovicZ13,Tuan:tom:2019,DBLP:journals/sigpro/ChenTCTVGG13,Toan:wacv:2019,DBLP:journals/spm/GirodCCCGRTTV11}, place recognition~\cite{247,netvlad,Torii-PAMI2015}, camera pose estimation~\cite{Sattler1,Irschara,man-tip19}. State-of-the-art image search systems \cite{DBLP:conf/cvpr/JegouZ14,DBLP:conf/cvpr/ArandjelovicZ13,herve_cvpr2010,DBLP:conf/cvpr/ArandjelovicZ12,F-FAemb} include three main steps in computing the image representation: local feature extraction, embedding, and aggregating. 
The local feature extraction step extracts a set of local features, e.g. SIFT \cite{SIFT_Lowe}, representing the image. The embedding step improves the discriminativeness of the local features by mapping these features into a high-dimensional space \cite{do_cvpr15,herve_cvpr2010,DBLP:conf/cvpr/JegouZ14,F-FAemb}. The aggregating (pooling) step converts the set of mapped high dimensional vectors into a single vector representation which usually has the dimensionality of several thousands \cite{do_cvpr15,herve_cvpr2010,DBLP:conf/cvpr/JegouZ14,F-FAemb}. In particular, the aggregating step is very important.  First, the aggregating step 
reduces the storage requirement which is one of main concerns in large-scale image search. Second, the aggregated representation vectors enables direct comparison  using standard metrics such as Euclidean distance.

Although the aggregated representation reduces the storage and allows simple distance-based comparison, it is not efficient enough for large-scale database which requires very compact representation and fast searching. An attractive approach for achieving these requirements is binary hashing. Specifically, binary hashing encodes image representations into compact binary hash codes, in which distances among data points can be efficiently calculated using  bit operations, i.e., XOR and POPCOUNT. Furthermore,  binary representations reduce storage significantly. 

Although both aggregating and hashing play important roles in large scale image search systems, in most works, the aggregating and hashing steps are designed  independently and separately \cite{DBLP:conf/cvpr/GongKRL13,DBLP:conf/cvpr/KimC15,DBLP:journals/pami/HeoLHCY15}:
First, some aggregation is applied on the local (embedded) features, resulting in a single aggregated representation for each image. Then, the set of aggregated representations is used for learning a hash function which encodes the aggregated representations into compact binary codes.  For example,   
Generalized Max Pooling \cite{gmp}
seeks a representation that can achieve some desirable aggregation property, i.e., equalizing the similarity between the representation and individual local features.
This aggregation process does not take into account any aspect of the subsequent hashing, and the resulted representations may not be optimal for hashing, e.g., in the context of unsupervised hashing, the aggregated representation may be difficult to be reconstructed by binary codes.
\red{In this work, we propose a novel framework where feature aggregating and hashing are designed simultaneously and optimized jointly. Specifically, in our proposed framework, we aim to compute aggregated representations that not only can achieve some desired aggregation property (equalized similarity) but also can be better reconstructed by some binary codes (in the unsupervised setting) or can better preserve the semantic similarity (in the supervised setting). As the aggregation is more reconstructible (for the unsupervised setting) and can preserve more semantic information (for the supervised setting), the binary codes are discriminative, resulting in improved retrieval performance. 
}

{\bf Our specific contributions are:} 
\textit{(i)} In order to accelerate simultaneous learning of aggregating and hashing, we first propose a relaxed version of the state-of-the-art unsupervised hashing {Binary Autoencoder} \cite{BA_CVPR15} to be used in our framework. Instead of solving a NP-hard problem with the hard binary constraint on the outputs of the encoder, we propose to solve the problem with relaxation of the binary constraint, i.e., minimizing the binary quantization loss. In order to minimize this loss, we propose to solve the problem with alternating optimization. This proposed hashing method is not only faster in training but also competitive in retrieval accuracy in comparison to Binary Autoencoder \cite{BA_CVPR15}.
\textit{(ii)} Our second contribution is a simultaneous feature aggregating and hashing learning approach which takes a set of local (embedded) features\footnote{In this work, the embedding is applied when SIFT features are used.} as the input and learns the aggregation and the hash  function simultaneously.
We propose alternating optimization for learning the aggregated features and the hash function. The proposed relaxed Binary Autoencoder is used for learning the hash parameters in the alternating optimization. 
\textit{(iii)} we extend the framework to supervised hashing by leveraging the label information such that the binary codes preserve the semantic similarity of samples. Furthermore, in the supervised setting, the binary codes of a new testing image can not be directly computed from learned model parameters because the aggregating process requires image label which is not available for the testing image. To overcome this challenge, we propose a novel simple yet powerful solution that learns a mapping from original aggregated features to learned aggregated features. The learned mapping is then used in the process of computing binary codes for new images. 
\textit{(iv)} We perform solid experiments on different image retrieval benchmark datasets to evaluate the proposed frameworks. We also evaluate the proposed methods with different state-of-the-art  image features under different configurations. The experimental results show that the proposed simultaneous learning outperforms other recent unsupervised and supervised hashing methods.

 A preliminary version of this work  has been presented in~\cite{DBLP:conf/cvpr/DoTPC17}. The extensions in this current version are:  Firstly, the introduction and the related work sections are fully revised to clearly describe our contributions and to thoroughly cover recent works. Secondly, we provide extensive comparisons between the proposed unsupervised framework to recent state-of-the-art deep learning-based unsupervised hashing methods.  Thirdly, we adapt the proposed unsupervised framework to supervised hashing. In particular, we reformulate problem formulations and optimization processes by using the label information to supervise the learning. The binary codes are learned such that they not only encourage the aggregating property but also optimize for a linear classifier.  We also propose a novel simple yet powerful solution to compute the binary codes of testing images. In addition, we also provide the asymptotic complexity the proposed method. Finally, we extensively evaluate and compare the proposed supervised framework to both state-of-the-art non-deep-based and deep-based supervised hashing methods under various configurations, i.e., traditional setting and unseen class setting. The experimental results show that our supervised method outperforms the state-of-the-art supervised hashing methods.

The remaining of this paper is organized as follows. Section \ref{sec:relatedwork} presents related works. Section \ref{sec:RBA} introduces the Relaxed Binary Autoencoder. Sections \ref{sec:SAH} and \ref{sec:evaSAH} introduce and evaluate the proposed simultaneous feature aggregating and hashing (SAH) for unsupervised hashing, respectively. Sections \ref{sec:SASH} and \ref{sec:evaSASH} introduce and evaluate the proposed simultaneous feature aggregating and supervised hashing (SASH), respectively. Section \ref{sec:concl} concludes the paper.

\section{Related work}
\label{sec:relatedwork}
\begin{table}[!t]
\footnotesize
\centering
\caption{Notations and their corresponding meanings.}% in the description of the method.}
\label{tab:notation}
\begin{tabular}{|l|l|}
\hline
Notation	&Meaning \\ \hline
$\X$ &$\X = \{\x_i\}_{i=1}^{m} \in \R^{D\times m}$: set of $m$ training samples; \\
	 &each column of $\X$ corresponds to one sample\\ \hline	
$\Y$ &$\Y = \{\y_i\}_{i=1}^m \in \R^{C \times m}$: set of label vectors\\
	 &for supervised setting\\ \hline	
$\B$ &$\B = \{\bb_i\}_{i=1}^{m} \in \{-1,+1\}^{L\times m}$: binary code matrix \\ \hline
$L$  &Number of bits to encode a sample \\ \hline		
$\W_1, \cc_1$&$\W_1 \in \R^{L\times D}, \cc_1 \in \R^{L\times 1}$: weight and bias of encoder \\ \hline
%$\W_2, \cc_2$&$\W_2 \in \R^{D\times L}, \cc_2 \in \R^{D\times 1}$: weight and bias of decoder \\ \hline

$\W_2, \cc_2$& weight and bias of decoder. \\
&$\bullet$ $\W_2 \in \R^{D\times L}, \cc_2 \in \R^{D\times 1}$: unsupervised setting\\
& $\bullet$ $\W_2 \in \R^{C\times L}, \cc_2 \in \R^{C\times 1}$: supervised setting\\
\hline
$\Vs$ &$\Vs=\{\V_i\}_{i=1}^m$; $\V_i\in \R^{D\times n_i}$ is set of local (embedded)\\ 
	 &representations of image $i$; \\
	 &$n_i$ is number of local descriptors of image $i$\\ \hline%local (embedded) image representations; \\
	 %&$\V_i\in \R^{}$ corresponds to local representations of image $i$ \\ \hline	
$\Phi$ &$\Phi = \{\varphi_i\}_{i=1}^{m} \in \R^{D\times m}$: set of $m$ aggregated vectors; \\
	 &$\varphi_i$ corresponds to aggregated vector of image $i$\\ \hline	
$\1$ & column vector with all $1s$ elements\\ \hline
$\I$ & identity matrix \\ \hline
\end{tabular}
\end{table}
In order to make the paper clear and easy to follow, we summarize the used notations in Table~\ref{tab:notation}. Two main components of the proposed simultaneous learning are aggregating and hashing. For aggregating, we rely on the state-of-the-art {Generalized Max Pooling} \cite{gmp}. For hashing, we propose a relaxed version of Binary Autoencoder \cite{BA_CVPR15}. %This section presents a brief overview of Generalized Max Pooling \cite {gmp} and Binary Autoencoder \cite{BA_CVPR15}.  
This section presents a brief overview of Generalized Max Pooling \cite {gmp} and hashing methods. %Specifically, we will focus on the  most two related hashing methods, i.e., Iterative Quantization~\cite{DBLP:conf/cvpr/GongL11} and Binary Autoencoder \cite{BA_CVPR15}.  

\subsection{Generalized Max Pooling (GMP) \cite{gmp}}
Sum-pooling~\cite{DBLP:conf/cvpr/LazebnikSP06} and max-pooling~\cite{DBLP:conf/cvpr/YangYGH09,DBLP:conf/icml/BoureauPL10} are two common methods for aggregating set of local (embedded) vectors of an image to a single vector. However, sum-pooling lacks discriminability because the aggregated vector is more influenced by frequently-occurring uninformative descriptors than rarely-occurring informative ones. Max-pooling equalizes the influence of frequent and rare descriptors. However, classical max-pooling approaches can only be applied to BoW or sparse coding features. 
%Max-pooling~\cite{DBLP:conf/cvpr/YangYGH09,DBLP:conf/icml/BoureauPL10} is a common aggregation method which aggregates a set of local (embedded) vectors of the image to a single vector. However, classical max-pooling approach can only be applied to BoW or sparse coding features. 
To overcome this challenge, in \cite{DBLP:conf/cvpr/JegouZ14} and \cite{gmp} the authors concurrently introduced a generalization of max-pooling (i.e., Generalized Max Pooling (GMP) \cite{gmp})\footnote{In \cite{DBLP:conf/cvpr/JegouZ14}, the authors named their method as \textit{democratic aggregation}. It actually shares similar idea to \textit{generalized max pooling} \cite{gmp}} that can be applied to general features such as VLAD \cite{jegou11d}, Temb \cite{DBLP:conf/cvpr/JegouZ14}, Fisher vector \cite{DBLP:conf/cvpr/PerronninD07}. The main idea of GMP is to equalize the similarity between each local embedded vector and the aggregated representation. In \cite{DBLP:conf/cvpr/JegouZ14,F-FAemb}, the authors  showed that GMP achieves better retrieval accuracy than sum-pooling.
%man: pls confirm dimension 

Given $\V \in \R^{D\times n}$, the set of $n$ embedded vectors of an image (each embedded vector has dimensionality $D$), GMP finds the aggregated representation $\varphi$ which equalizes the similarity (i.e. the dot-product)
 between each column of $\V$ and $\varphi$ by solving the following optimization
\begin{equation}
\min_{\varphi}\left(\norm{\V^T\varphi-\1}^2+\mu\norm{\varphi}^2 \right) \label{eq:gmp}
\end{equation}
(\ref{eq:gmp}) is a ridge regression problem which solution is %is a $l_2$ regularized least-squares and its analytic solution is 
\begin{equation}
\varphi = \left( \V\V^T + \mu\I\right)^{-1} \V\1 \label{eq:gmp-solution}
\end{equation}

%\subsection{Binary Autoencoder (BA)\cite{BA_CVPR15}} 
\subsection{Hashing methods}
Existing binary hashing methods can be categorized as data-independent and data-dependent schemes \cite{DBLP:journals/corr/WangLKC15,DBLP:journals/corr/WangSSJ14_journal,Grauman_review}. 
Data-independent hashing methods~\cite{lsh_vldb09,KLSH_iccv09,KLSH_nips09,DBLP:journals/pami/KulisJG09} rely on random projections for constructing hash functions. Although representative data-independent hashing methods such as Locality-Sensitive Hashing (LSH)~\cite{lsh_vldb09} and its kernelized versions~\cite{KLSH_iccv09,KLSH_nips09} have theoretical guarantees that the more similar data would have higher probability to be mapped into similar binary codes, they require long codes to achieve high precision. 
Different from data-independent approaches, data-dependent hashing methods use available training data for learning hash functions in unsupervised or supervised manner and they usually achieve better retrieval results than data-independent methods. The unsupervised hashing methods~\cite{DBLP:conf/nips/WeissTF08,DBLP:conf/cvpr/GongL11,DBLP:conf/cvpr/HeWS13,BA_CVPR15,CVPR12:SphericalHashing,DeepHash_TIP17} try to preserve the neighbor similarity of samples in Hamming space without label information. The representative unsupervised hashing methods are Iterative Quantization (ITQ)~\cite{DBLP:conf/cvpr/GongL11}, Spherical Hashing (SPH)~\cite{CVPR12:SphericalHashing}, K-means Hashing (KMH)~\cite{DBLP:conf/cvpr/HeWS13}, etc. 
The supervised hashing methods~\cite{Kulis_learningto,CVPR12:Hashing,CVPR2014Lin,Shen_2015_CVPR,DBLP:conf/icml/NorouziF11} try to preserve the label similarity of samples using labeled training data. The representative supervised hashing methods are Kernel-Based Supervised Hashing (KSH)~\cite{CVPR12:Hashing}, Semi-supervised Hashing (SSH)~\cite{DBLP:journals/pami/WangKC12}, Supervised Discrete Hashing (SDH)~\cite{Shen_2015_CVPR}, 
\red{Asymmetric Inner-product Binary Coding (AIBC) \cite{AIBC}, Graph Convolutional Network Hashing (GCNH) \cite{GCNH},} etc.  

Most of the previous hashing methods are originally designed and experimented on hand-crafted features which may limit their performance in practical applications. Recently, to leverage the power of deep convolutional neural networks (CNNs)~\cite{Lecun98gradient-basedlearning,krizhevsky2012imagenet,Simonyan14c}, many deep unsupervised and supervised hashing methods have been proposed. In~\cite{DBLP:conf/aaai/XiaPLLY14} the authors proposed a two-step supervised hashing method which learns a deep CNN based hash function with the pre-computed binary codes. In~\cite{DBLP:conf/cvpr/LaiPLY15,DBLP:conf/aaai/ZhuL0C16,DBLP:conf/aaai/CaoL0ZW16,DBLP:conf/cvpr/ZhaoHWT15,DBLP:journals/tip/ZhangLZZZ15,DSDH} the authors proposed end-to-end deep supervised hashing methods in which the image features and the hash codes are simultaneously learned. Most of those models consist of a deep CNN for image feature extraction and a binary quantization component that tries to approximate the \textit{sgn} function. 
Different from supervised setting, there are few end-to-end hashing which are proposed for unsupervised setting.  In \cite{deepbit2016}, the authors proposed an end-to-end deep learning  framework for unsupervised hashing. The network is trained to produce hash codes that minimize the quantization loss w.r.t. the output of the last VGG's~\cite{Simonyan14c} fully connected layer. Recently, in~\cite{SADH} the authors proposed an unsupervised deep hashing method that alternatingly proceeds over three training modules: deep hash model training, similarity graph updating and binary code optimization.
Different from previous deep hashing methods that try to simultaneously learn image features and  binary codes, our work uses a pre-trained deep model to extract local image representations. These representations are used as inputs for the simultaneous learning of the aggregated representation and the binary codes. 

One of problems which makes the binary hashing difficult is the binary constraint on the codes, i.e., the outputs of the hash functions have to be binary. Generally, this binary constraint leads to an NP-hard mixed-integer optimization problem. In order to overcome this difficulty, several approaches have been proposed in the literature.  
In Deep Hashing (DH) \cite{DeepHash_TIP17}, the binary constraint is handled by applying the $sign$ function on the outputs of the last layer of the network. The authors assumed that the $sign$ function is differentiable everywhere during training. In Iterative Quantization  \cite{DBLP:conf/cvpr/GongL11}, the binary constraint is relaxed by minimizing the binary quantization loss. In Binary Autoencoder \cite{BA_CVPR15}  the binary constraint is handled by using the $sign$ function. In order to overcome the non-differentiable of the $sign$ function, the authors \cite{BA_CVPR15} use binary SVMs to learn the model parameters.
In \cite{DBLP:conf/mm/ZhangWHC16}, the authors also used the $sign$ function to handle the binary constraint. To deal with the non-differentiable problem of the sign function, the authors proposed to use the hinge loss to approximate the sign function. 
In \cite{do2016learning}, to handle the binary constraint, the authors relied on the idea of minimizing the binary quantization error which is achieved by an alternating optimization over the real-valued network weights and auxiliary binary variables.

\red{There is another research topic that is related to binary hashing, i.e., training neural networks with binary weights. In this problem, the network weights are constrained to be binary. In the recent work, Binary Connect \cite{bin_connect}, the authors use a stochastic binarization function to binarize the network weights. 
To handle the gradient problem of the binarization function which is zero almost everywhere, the authors utilized the straight-through estimator \cite{STE} to approximate the gradient of the  function. It is worth noting that the activation outputs of the Binary Connect \cite{bin_connect}  are still real-valued, which is different from hashing problem which aims to produce binary outputs.}
In the following, we brief the two most related hashing methods to our work, i.e., Iterative Quantization (ITQ)~\cite{DBLP:conf/cvpr/GongL11} and Binary Autoencoder (BA) \cite{BA_CVPR15}.

\textbf{Iterative Quantization (ITQ)~\cite{DBLP:conf/cvpr/GongL11}: }
In~\cite{DBLP:conf/cvpr/GongL11}, the authors propose ITQ which is a two-step hashing method. In the first step, ITQ computes continuous low-dimensional codes by applying PCA to the data. In the second step, it finds a rotation that makes the PCA codes as close as possible to binary values. The second step is actually  an Orthogonal Procrustes problem~\cite{Schonemann1966}  in which one tries to find a rotation to align one point set with another and it has a closed-form solution by using SVD. 
ITQ is a postprocessing of the PCA codes and it can be seen as a suboptimal approach to optimizing a binary autoencoder, where the binary constraints
are relaxed during the optimization (i.e., in the PCA step), and then one projects the continuous codes back to the binary space. 

\textbf{Binary Autoencoder (BA)\cite{BA_CVPR15}:}
In \cite{BA_CVPR15}, instead of ignoring binary constraints during the dimensionality reduction and then binarizing the continuous codes, the authors propose a joint optimization.  Specifically, in order to compute the binary codes, the authors minimize the following optimization
%\vspace{-0.3cm}
\begin{equation}
\min_{\h,\f,\Z} \sum_{i=1}^m\left(\norm{\x_i - \f(\z_i)}^2+\mu\norm{\z_i-\h(\x_i)}^2 \right)
\label{eq:ba}
\end{equation}
\vspace{-0.2cm}
\begin{equation}
\textrm{s.t.\ } \z_i \in \{-1,1\}^{L}, i = 1,...,m \label{eq:ba-constraint}
\end{equation}
where $\h=sgn(\W_1\x + \cc_1)$ and $\f$ are encoder and decoder, respectively. By having $sgn$, the encoder will output binary codes. In the training of BA, the authors compute each variable $\f,\h,\Z$ at a time while holding the other fixed. 
The authors show that the BA outperforms state-of-the-art unsupervised  hashing methods. 
However, the disadvantage of BA is the time-consuming training which is mainly caused by the computing of $\h$ and $\Z$. As $\h$ involves $sgn$, it cannot be solved analytically. Hence, when computing $\h$, the authors cast the problem as the learning of $L$ separated linear SVM classifiers, i.e., for each $l=1,...,L$, they fit a linear SVM to $(\X, \Z_{l,.})$. When computing $\Z$, the authors solve for each sample $\x_i$ independently. 
Solving $\z_i$ in (\ref{eq:ba}) 
for each sample under the binary constraint (\ref{eq:ba-constraint}) is NP-hard. To handle this, the authors first solve the problem with the relaxed constraint $\z_i \in [-1,1]$, resulting a continuous solution. 
They then apply the following procedure several times for getting $\z_i$: 
for each bit from $1$ to $L$, they evaluate the objective function with the bit equals to $-1$ or $1$ with all remaining elements fixed and pick the best value for that bit. The asymptotic complexity for computing $\Z$ over all samples is $\mathcal{O}(mL^3)$.

In the following, we introduce our efficient Relaxed Binary Autoencoder algorithm (Section \ref{sec:RBA}) which will be used in the proposed simultaneous feature aggregating and hashing framework (Section \ref{sec:SAH}). 

\section{Relaxed Binary Autoencoder (RBA)}
\label{sec:RBA}
\subsection{Formulation}
In order to achieve binary codes, we propose to solve the following constrained optimization

\vspace{-0.3cm}\footnotesize
\begin{eqnarray}
\min_{\{\W_i,\cc_i\}_{i=1}^{2}} J &=& \frac{1}{2} \norm{\X-\left(\W_2(\W_1\X+\cc_1\1^T)+\cc_2\1^T\right)}^2 \nonumber \\ 
{}&&+\frac{\beta}{2}\left(\norm{\W_1}^2+\norm{\W_2}^2\right) \label{eq:obj_ori}
\end{eqnarray}
\begin{equation}
\textrm{s.t. } \W_1\X+\cc_1\1^T \in \{-1,1\}^{L\times m} \label{eq:binary0}
\end{equation} 
\normalsize 
The constraint (\ref{eq:binary0}) makes sure the output of the encoder is binary. The first term of (\ref{eq:obj_ori}) makes sure the binary codes provide a good reconstruction of the input, so it encourages (dis)similar inputs map to (dis)similar binary codes. The second term is a regularization that tends to decrease the magnitude of the weights, so it helps to prevent overfitting.

Solving (\ref{eq:obj_ori}) under (\ref{eq:binary0}) is difficult due to the binary constraint. In order to overcome this challenge, we propose to solve the relaxed version of the binary constraint, i.e., minimizing the binary quantization loss of the encoder. The proposed method is named as \textit{Relaxed Binary Autoencoder} (RBA). Specifically, inspired from the quadratic penalty method for constrained optimization~\cite{Nocedal06}, 
we introduce a new auxiliary variable $\B$ and solve the following the optimization 

% \vspace{-0.3cm}
\footnotesize
\begin{eqnarray}
\min_{\{\W_i,\cc_i\}_{i=1}^{2},\B} J &=& \frac{1}{2} \norm{\X-\left(\W_2\B+\cc_2\1^T\right)}^2 \nonumber \\ 
{}&&\hspace{-8em}+\frac{\lambda}{2}\norm{\B-(\W_1\X+\cc_1\1^T)}^2+\frac{\beta}{2}\left(\norm{\W_1}^2+\norm{\W_2}^2\right) \label{eq:obj_2}
\end{eqnarray}
\begin{equation}
\textrm{s.t. } \B \in \{-1,1\}^{L\times m} \label{eq:binary2}
\end{equation} 
\normalsize 
The benefit of the auxiliary variable $\B$ is that we can decompose the difficult constrained optimization problem (\ref{eq:obj_ori}) into simpler sub-problems.  We use alternating optimization on these sub-problems as will be discussed in detail.

%i.e., optimization over $\W_i$, $\cc_i$ and $\B$. Hence, we can alternative optimize with respect to each sub-problem while holding the other fixed. 

%The proposed RBA ((\ref{eq:obj_2}) and (\ref{eq:binary2})) has similar form as BA ((\ref{eq:ba}) and (\ref{eq:ba-constraint})). 
An important difference between the proposed RBA and the original BA is that our encoder does not involve $sgn$ function. The second term of (\ref{eq:obj_2}) forces the output of encoder close to binary values, i.e., it minimizes the binary quantization loss, while the first term still ensures good reconstruction loss. %The parameter $\lambda$ serves a trade-off between two kinds of loss. 
By setting the penalty parameter $\lambda$ sufficiently large, we penalize the binary constraint violation severely, thereby forcing the solution of (\ref{eq:obj_2}) closer to the feasible region of the original problem (\ref{eq:obj_ori}).

\subsection{Optimization}
\label{subsec:RBA_opt}
In order to solve for $\W_1, \cc_1, \W_2, \cc_2$, $\B$ in (\ref{eq:obj_2}) under constraint (\ref{eq:binary2}), we solve each variable at a time while holding the other fixed.

\textbf{$(\W,\cc)$-step:} When fixing $\cc_1,\cc_2$ and $\B$, we have the closed forms for $\W_1, \W_2$ as follows
%By taking the derivative of (\ref{eq:obj_2}) w.r.t. each variable and assign to zero, we have the closed-form solution for $\W_1, \cc_1, \W_2, \cc_2$ as follow

\vspace{-0.2cm}\small
\begin{equation}
\W_1 = \lambda \left(\B-\cc_1\1^T\right)\X^T \left(\lambda\X\X^T+\beta\I\right)^{-1}
\label{eq:W1}
\end{equation}
\begin{equation}
\W_2 = \left(\X-\cc_2\1^T\right)\B^T \left(\B\B^T+\beta\I\right)^{-1}
\label{eq:W2}
\end{equation}
\normalsize 
When fixing $\W_1,\W_2$ and $\B$, we have the closed forms for $\cc_1, \cc_2$ as follows
\vspace{-0.2cm}
\begin{equation}
\cc_1 = \frac{1}{m}\left(\B-\W_1\X\right)\1
\label{eq:c1}
\end{equation}
\begin{equation}
\cc_2 = \frac{1}{m}\left(\X-\W_2\B\right)\1
\label{eq:c2}
\end{equation}
\normalsize
Note that in (\ref{eq:W1}), the term $\X^T \left(\lambda\X\X^T+\beta\I\right)^{-1}$ is a constant matrix and it is computed only one time. 

%\vspace{-0.3cm}
\textbf{$\B$-step:} When fixing the weight and the bias, by defining $\widetilde{\X}$ and $\H$ as follows
\begin{equation}
\widetilde{\X}=\X-\cc_2\1^T
\end{equation}
\begin{equation}
\H=\W_1\X+\cc_1\1^T
\end{equation}
we can rewrite (\ref{eq:obj_2}) as
\begin{equation}
\norm{\widetilde{\X}-\W_2\B}^2 + \lambda \norm{\H-\B}^2
\end{equation}
\begin{equation}
\textrm{s.t. } \B \in \{-1,1\}^{L\times m} 
\end{equation} 

Inspired by the recent progress of discrete optimization \cite{Shen_2015_CVPR}, 
we use coordinate descent approach for solving $\B$, i.e., we solve one row of $\B$ each time while fixing all other rows. Specifically, let $\Q = \W_2^T\widetilde{\X} + \lambda \H$; for $k=1,...,L$, let $\w_k$ be $k^{th}$ column of $\W_2$; $\overbar{\W}_2$ be matrix $\W_2$ excluding $\w_k$; $\q_k$ be $k^{th}$ column of $\Q^T$; $\bb_k^T$ be $k^{th}$ row of $\B$; $\overbar{\B}$ be matrix $\B$ excluding $\bb_k^T$. We have the closed-form solution for $\bb_k^T$ as
\begin{equation}
\bb_k^T  = sgn \left(\q_k^T - \w_k^T\overbar{\W}_2\overbar{\B}\right)
\label{eq:bk}
\end{equation}
The proposed RBA is summarized in Algorithm~\ref{alg1}. In the Algorithm~\ref{alg1}, $\B^{(t)}$, $\W_1^{(t)}, \cc_1^{(t)}, \W_2^{(t)}, \cc_2^{(t)}$ are values at $t^{th}$ iteration. 
After learning ($\W_1, \cc_1, \W_2, \cc_2$), given a new vector $\x$, we pass $\x$ to the encoder, i.e., $h = \W_1\x+\cc_1$, and round the values of $h$ to $\{-1,1\}$, resulting binary codes. 
\begin{algorithm}[!t]
	\footnotesize
	\caption{Relaxed Binary Autoencoder (RBA)}
	\begin{algorithmic}[1] 
		\Require 
			\Statex $\X$: training data; $L$: code length; $T_1$: maximum iteration number; parameters $\lambda, \beta$
		\Ensure 
			\Statex 
			Parameters $\W_1, \cc_1, \W_2, \cc_2$
			\Statex 
			\State Initialize $\B^{(0)} \in \{−1,1\}^{L\times m}$ using ITQ~\cite{DBLP:conf/cvpr/GongL11}
			\State Initialize $\cc_1^{(0)}=\mathbf{0}$, $\cc_2^{(0)}=\mathbf{0}$ 
			\For{$t = 1 \to T_1$}
				\State Fix $\B^{(t-1)}, \cc_1^{(t-1)}, \cc_2^{(t-1)}$, solve $\W_1^{(t)}, \W_2^{(t)}$ %by (\ref{eq:W1}) and (\ref{eq:W2}).
				\State Fix $\B^{(t-1)}, \W_1^{(t)}, \W_2^{(t)}$, solve $\cc_1^{(t)}, \cc_2^{(t)}$ %by (\ref{eq:c1}) and (\ref{eq:c2}).
				\State Fix $\W_1^{(t)}, \W_2^{(t)}, \cc_1^{(t)}, \cc_2^{(t)}$, solve $\B^{(t)}$ by \textbf{B-step}
			\EndFor
			\State Return %$\B_{(max\_iter)}$ and 
			$\W_1^{(T_1)}, \W_2^{(T_1)}, \cc_1^{(T_1)}, \cc_2^{(T_1)}$
    \end{algorithmic}
    \label{alg1}
\end{algorithm}

%\vspace{-0.3cm}
\textbf{Comparison to Binary Autoencoder (BA) \cite{BA_CVPR15}:} There are two main advances of the proposed RBA (\ref{eq:obj_2}) over BA (\ref{eq:ba}). First, our encoder does not involve the $sgn$ function. Hence, during the iterative optimization, instead of using SVM for learning the encoder as in BA, we have an analytic solution ((\ref{eq:W1}) and (\ref{eq:c1})) for the encoder. Second, when solving for $\B$, %in the iterative optimization, 
instead of solving each sample at a time as in BA, we solve all samples at the same time by adapting the recent advance discrete optimization technique \cite{Shen_2015_CVPR}. 
The asymptotic complexity for computing one row of $\B$, i.e. (\ref{eq:bk}), is $\mathcal{O}(mL)$. Hence the asymptotic complexity for computing $\B$ is only $\mathcal{O}(mL^2)$ which is less than $\mathcal{O}(mL^3)$ of BA. These two advances make the training of RBA is faster than BA. 

\begin{figure}[!t]
\centering
%\vspace{-0.4cm}
\subfigure[CIFAR10]{
       \includegraphics[scale=0.27]{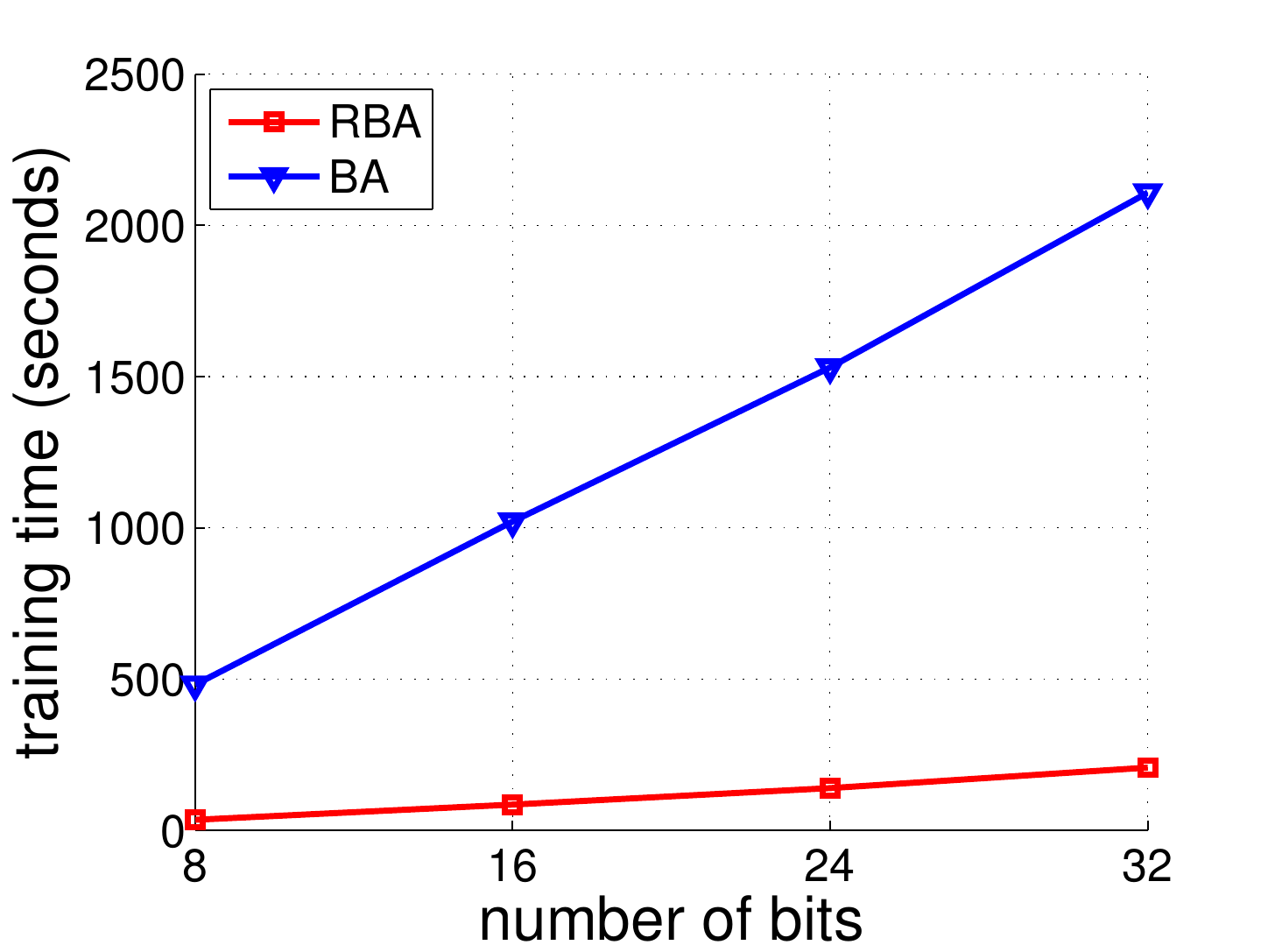}
       \label{fig:cifar_time}
}
% \hspace{-0.2cm}
\subfigure[SIFT1M]{
       \includegraphics[scale=0.27]{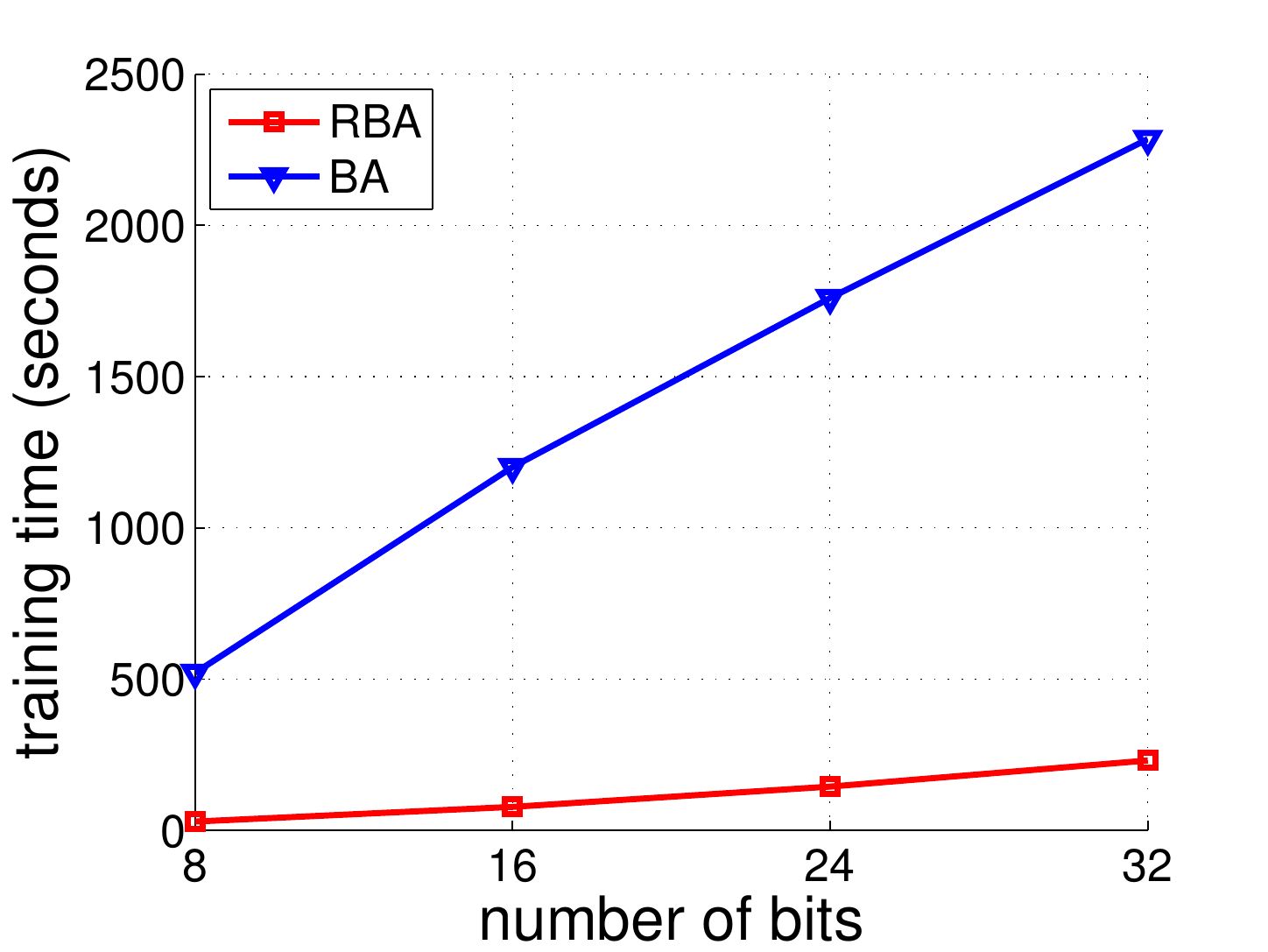} 
       \label{fig:sift_time}
}
\caption[]{Training time of BA and RBA on CIFAR10 and SIFT1M}
\label{fig:time-rba-ba}
\end{figure}
\begin{figure*}[!t]
\centering
\subfigure[CIFAR10]{
       \includegraphics[scale=0.33]{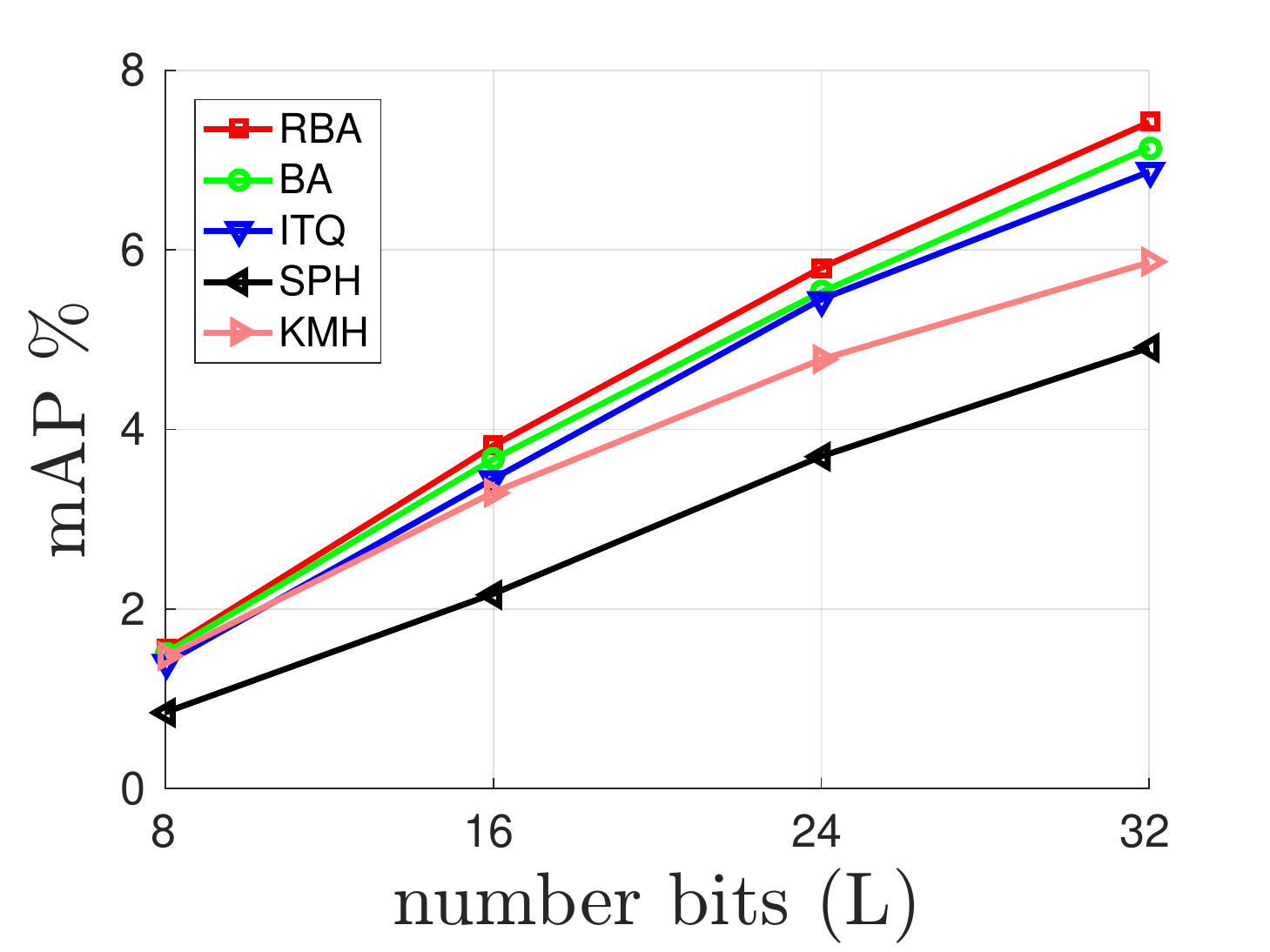}
       \label{fig:cifar_mAP}
}
\subfigure[MNIST]{
       \includegraphics[scale=0.33]{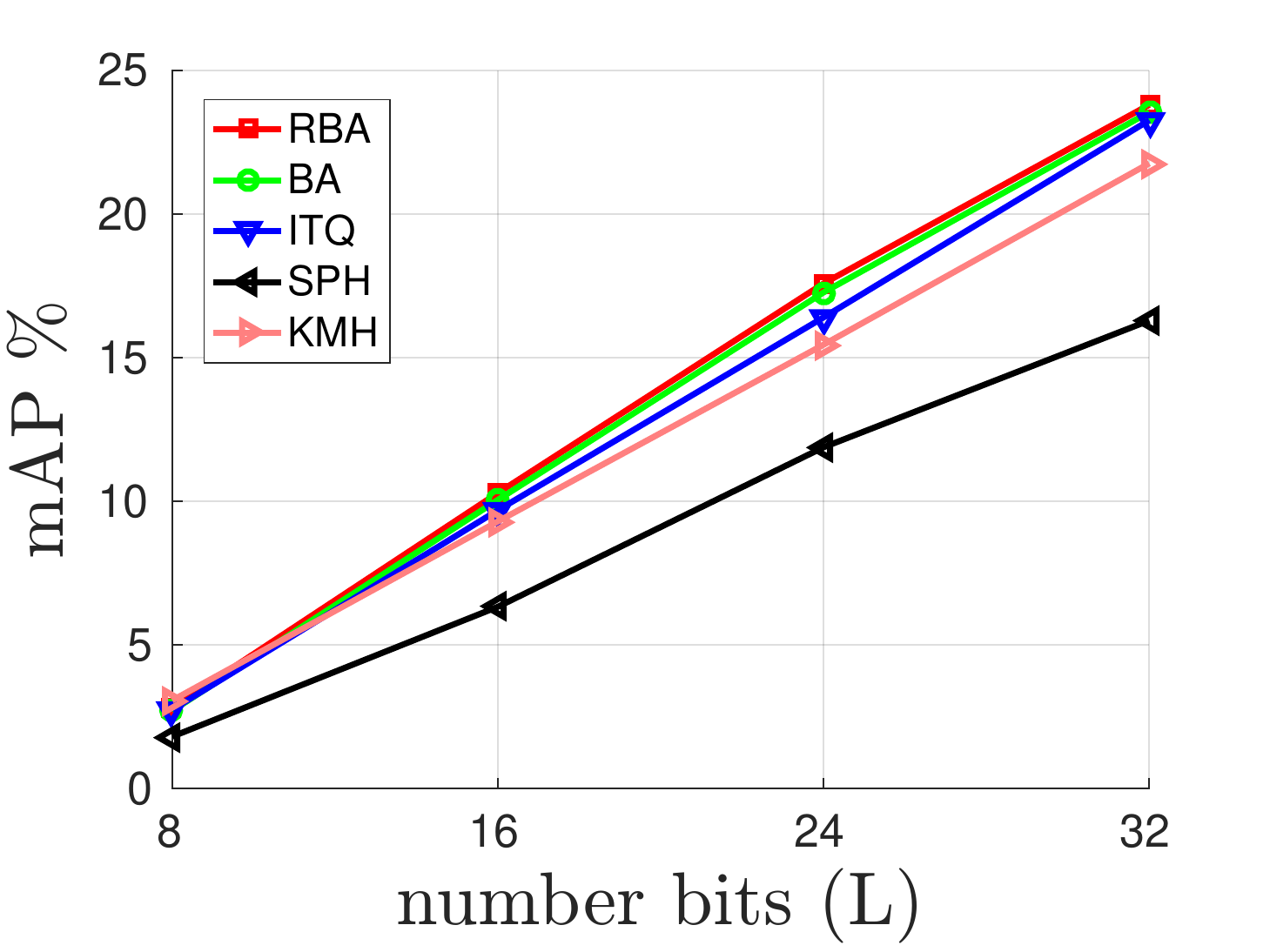} 
       \label{fig:mnist_mAP}
}
\subfigure[SIFT1M]{
       \includegraphics[scale=0.33]{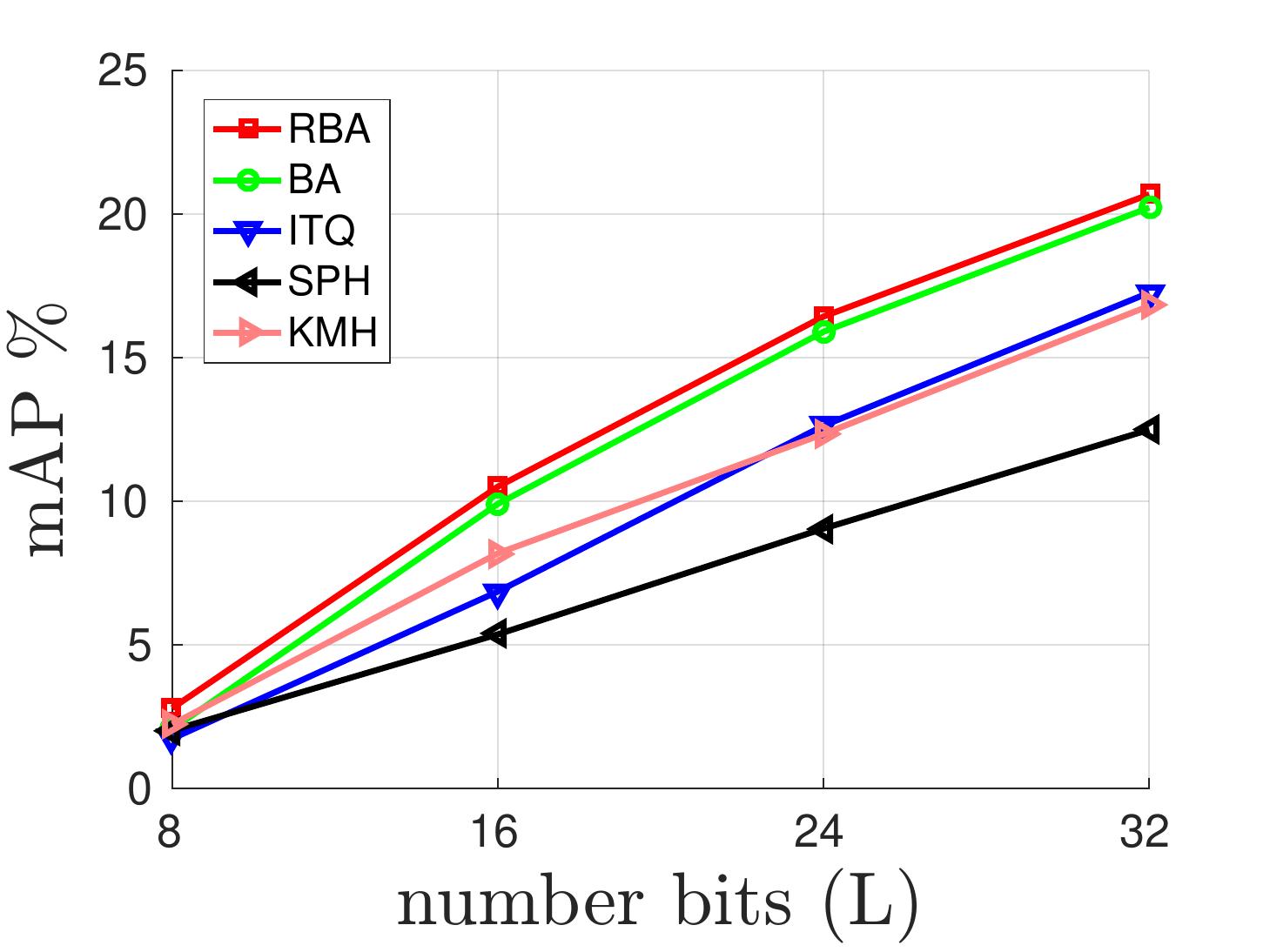} 
       \label{fig:sift1m_mAP}
}
\caption[]{mAP comparison between RBA and state-of-the-art unsupervised hashing methods on CIFAR10, MNIST, and SIFT1M}
\label{fig:rba-soa}
\end{figure*}
\subsection{Evaluation of Relaxed Binary Autoencoder (RBA)}
This section evaluates the proposed RBA and compares it to the following state-of-the-art unsupervised hashing methods:  
 Iterative Quantization (ITQ)~\cite{DBLP:conf/cvpr/GongL11}, Binary Autoencoder (BA)~\cite{BA_CVPR15}, K-means Hashing (KMH)~\cite{DBLP:conf/cvpr/HeWS13}, Spherical Hashing (SPH)~\cite{CVPR12:SphericalHashing}. For all compared methods, we use the implementations and the suggested parameters provided by the authors. The values of $\lambda$, $\beta$ and the number of iteration $T_1$ in the Algorithm \ref{alg1} are empirically set by cross validation as $10^{-2}, 1$ and $10$, respectively. 
 The BA \cite{BA_CVPR15} and the proposed RBA required an initialization for the binary code.  
 To make a fair comparison, 
 we follow \cite{BA_CVPR15}, i.e., using ITQ \cite{DBLP:conf/cvpr/GongL11} for the initialization.

\subsubsection{Dataset and evaluation protocol}
\label{subsub_eva_rba}
%\vspace{-0.2cm}
\textbf{\\}
\textbf{Dataset:} We conduct experiments on  CIFAR10 \cite{Krizhevsky09}, MNIST \cite{mnistlecun} and SIFT1M \cite{herve_pami2011} datasets which are widely used in evaluating hashing methods \cite{DBLP:conf/cvpr/GongL11,BA_CVPR15}. 

{\textbf{CIFAR10} dataset~\cite{Krizhevsky09}} consists of 60,000 images of 10 classes. 
The dataset is split into training and test sets, with $50,000$ and $10,000$ images,  respectively. Each image is represented by 320 dimensional GIST feature~\cite{gist}.

{\textbf{MNIST} dataset~\cite{mnistlecun}} consists of 70,000 handwritten digit
images of 10 classes. 
The dataset is split into training and test sets, with $60,000$ and $10,000$ images,  respectively. Each image is represented by a 784 dimensional gray-scale feature vector.% by using its intensity.

{\textbf{SIFT1M} dataset \cite{herve_pami2011}} contains 128 dimensional SIFT vectors \cite{SIFT_Lowe}. There are 1M vectors used as database for retrieval, 100K vectors for training, 
and 10K vectors for query. 
\\
%\vspace{-0.3cm}
\textbf{Evaluation protocol:}
In order to create ground truth for queries, we follow \cite{DBLP:conf/cvpr/GongL11,BA_CVPR15} in which the Euclidean nearest neighbors are used. The number of ground truths is set as in~\cite{BA_CVPR15}. For each query in CIFAR10 and MNIST datasets, its $50$ Euclidean nearest neighbors are used as ground truths; for each query in the large scale dataset SIFT1M, its 10,000  Euclidean nearest neighbors are used as ground truths. Follow the state of the art \cite{DBLP:conf/cvpr/GongL11,BA_CVPR15}, the performance of methods is measured by mAP. Note that as computing mAP is slow on the large scale dataset SIFT1M, we consider top 10,000 returned neighbors when computing mAP.
%\vspace{-0.2cm}
\subsubsection{Experimental results}
%\vspace{-0.2cm}
\textbf{\\Training time of RBA and BA:}
In this experiment, we empirically compare the training time of RBA and BA. The experiments are carried out on a processor core (Xeon E5-2600/2.60GHz). It is worth noting that the implementation of RBA is in Matlab, while BA optimizes the implementation by using mex-files %for learning SVMs 
at the encoder learning step. The comparative training time on CIFAR10 and SIFT1M datasets is showed in Figure \ref{fig:time-rba-ba}. The results show that RBA is more than ten times faster training than BA for all code lengths on both datasets. 
\\
\textbf{Retrieval results:}
Figure \ref{fig:rba-soa} shows the comparative mAP between methods. We find the following observations are consistent for all three datasets. At all code lengths, the proposed RBA outperforms or is competitive with the state-of-the-art BA. This result confirms the advance of our approach for computing encoder (i.e., closed-form) and \textbf{$\B$-step}  (i.e. using coordinate descent with closed-form for each row).
The results in Figure \ref{fig:rba-soa} also confirm the superior performance of BA and RBA over other methods. The improvements are more clear on the large scale SIFT1M dataset. 
\\
\red{
\textbf{Effects of hyper-parameters:} Figure \ref{fig:hyperparam} shows the retrieval performance of RBA with different $\beta$ and $\lambda$ values (in Eq.~\ref{eq:obj_2}) on CIFAR10 and SIFT1M datasets. We can observe that for CIFAR10, RBA generally achieves the best performance when $\beta\in [0.1, 1]$ and $\lambda\in[5\times 10^{-3}, 5\times 10^{-2}]$. While for SIFT1M, RBA is pretty robust to $\beta\in[0.1, 10]$ and $\lambda\in[0.001, 0.1]$.

\begin{figure}[t]
\centering
%\vspace{-0.4cm}
\subfigure[CIFAR10]{
       \includegraphics[scale=0.23]{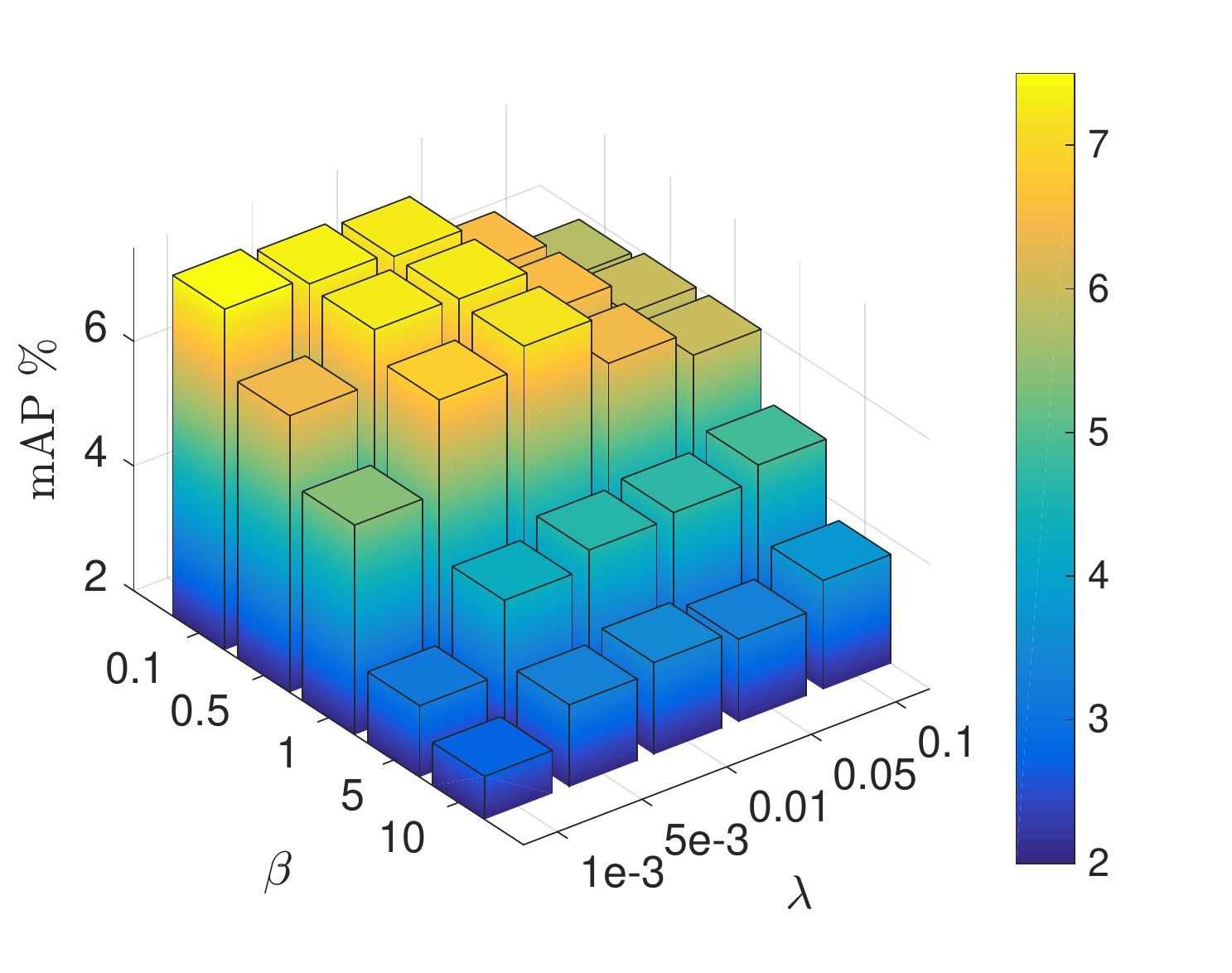}
       \label{fig:cifar_time}
}
% \hspace{-0.2cm}
\subfigure[SIFT1M]{
       \includegraphics[scale=0.23]{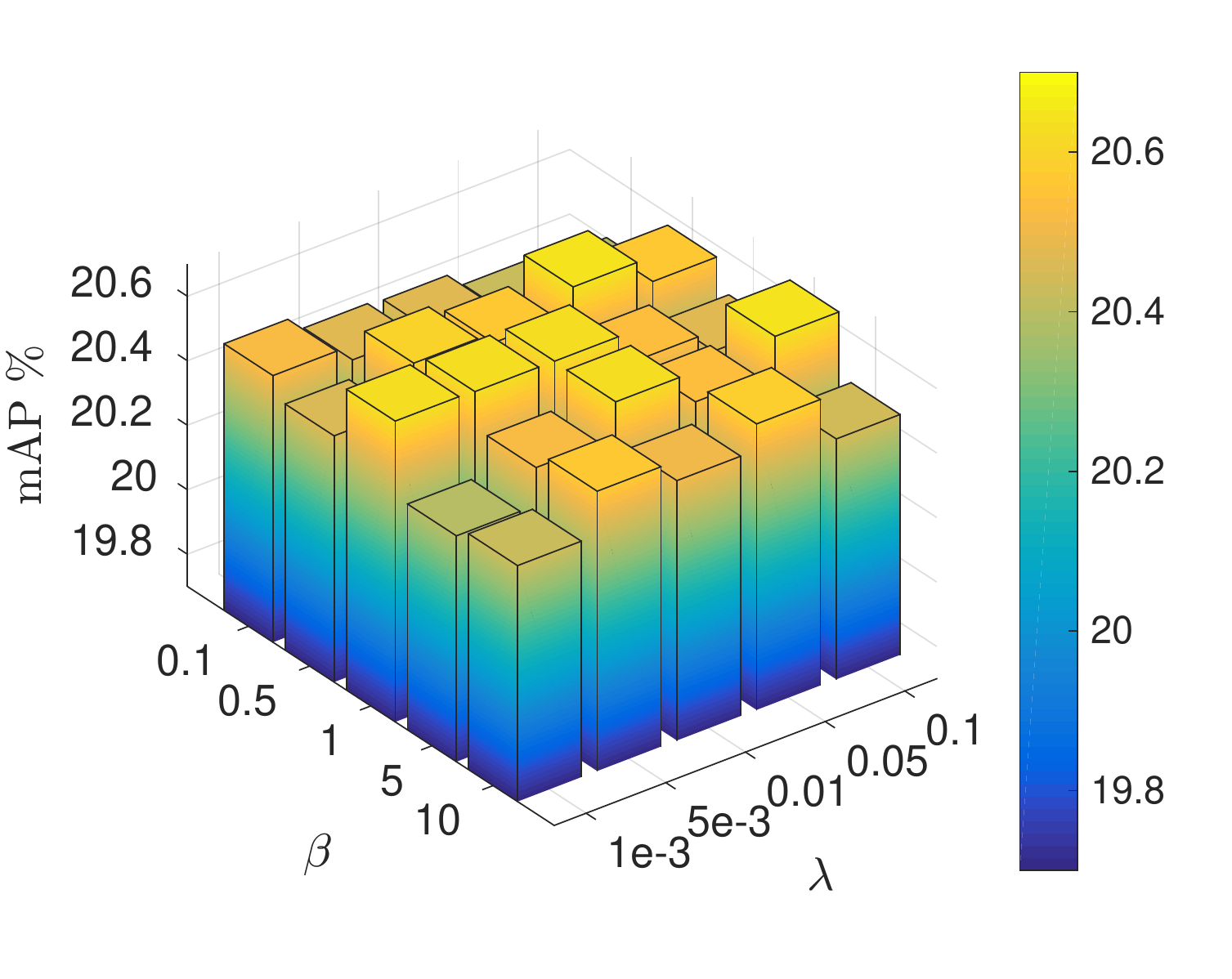} 
       \label{fig:sift_time}
}
\caption[]{The retrieval performance of RBA on CIFAR10 and SIFT1M for 32 bits when varying $\beta$ and $\lambda$.}
\label{fig:hyperparam}
\end{figure}

}

\section{Simultaneous Feature Aggregating and Hashing (SAH)}
\label{sec:SAH}
\subsection{Formulation}
Our goal is to simultaneously learn the aggregated vector representing an image and the hashing function, given the set of local image representations. 
For simultaneous learning, the learned aggregated vectors and the hash parameters should ensure desired properties of both aggregating and hashing.  Specifically, {\em aggregating property}: (i) for each image $i$, the dot-product similarity between the aggregated vector $\varphi_i$ and each local vector of $\V_i$ should be a constant; {\em hashing properties}: (ii) the outputs of the encoder are binary and (iii) the binary codes should preserve the similarity between image representations. In order to achieve these properties, we formulate the simultaneous learning as the following optimization

\vspace{-0.3cm}\footnotesize
\begin{eqnarray}
{}&&\hspace{-3em}\min_{\W_1,\cc_1,\W_2,\cc_2,\Phi} \frac{1}{2} \norm{\Phi-\left(\W_2(\W_1\Phi+\cc_1\1^T)+\cc_2\1^T\right)}^2 \nonumber \\ 
{}&&\hspace{-4em}+\frac{\beta}{2}\left(\norm{\W_1}^2+\norm{\W_2}^2\right)+\frac{\gamma}{2}\sum_{i=1}^{m}\left(\norm{\V_i^T\varphi_i-\1}^2+\mu\norm{\varphi_i}^2 \right) \label{eq:obj_ori_join}
\end{eqnarray}
\begin{equation}
\textrm{s.t. } \W_1\Phi+\cc_1\1^T \in \{-1,1\}^{L\times m} \label{eq:binary_join}
\end{equation} 
\normalsize 
The proposed constrained objective function (\ref{eq:obj_ori_join}) has a clear meaning. The first term of (\ref{eq:obj_ori_join}) ensures a good reconstruction of $\Phi$, hence it encourages the similarity preserving (the property iii). The binary constraint (\ref{eq:binary_join}) ensures the binary outputs of encoder (the property ii). Finally, the third term encourages the learned aggregated representation equals the similarities between $\varphi_i$ and different columns of $\V_i$ by forcing their inner product to be $1$ (the property i). 

\subsection{Optimization}
In order to solve (\ref{eq:obj_ori_join}) under constraint (\ref{eq:binary_join}), we propose to iteratively optimize it by alternatingly optimizing w.r.t. hashing parameters ($\W,\cc$) and  aggregated representation $\Phi$ while holding the other fixed. 

\textbf{$\Phi$-step:} When fixing $\W_1,\cc_1,\W_2,\cc_2$ and solving for $\Phi$, we can solve over each $\varphi_i$ independently. Specifically, for each sample $i=1,...,m$, we solve the following relaxed problem by skipping the binary constraint
\begin{eqnarray}
\min_{\varphi_i} \frac{1}{2} \norm{\varphi_i-\left(\W_2(\W_1\varphi_i+\cc_1)+\cc_2\right)}^2 \nonumber \\ 
{}&&\hspace{-17em}+\frac{\gamma}{2}\left(\norm{\V_i^T\varphi_i-\1}^2+\mu\norm{\varphi_i}^2 \right) \label{eq:obj_pool}
\end{eqnarray}
By solving (\ref{eq:obj_pool}), we find $\varphi_i$ which satisfies the properties (i) and (ii), i.e., $\varphi_i$ not only ensures the aggregating property but also minimize the reconstruction error w.r.t. the fixed hashing parameters. (\ref{eq:obj_pool}) is actually a $l_2$ regularized least squares problem, hence we achieve the analytic solution as

\vspace{-0.3cm}\footnotesize
\begin{eqnarray}
\varphi_i &=& \left((\I-\W_2\W_1)^T(\I-\W_2\W_1)+\gamma\V_i\V_i^T+\gamma\mu\I\right)^{-1} \nonumber \\ 
{}&&\hspace{0em}\times \left(\gamma\V_i\1+(\I-\W_2\W_1)^T(\W_2\cc_1+\cc_2)\right) \label{eq:sol_pool}
\end{eqnarray}
\normalsize

\textbf{$(\W,\cc)$-step:} When fixing $\Phi$ and solving for $(\W_1,\cc_1,\W_2,\cc_2)$, (\ref{eq:obj_ori_join}) under the constraint (\ref{eq:binary_join}) is equivalent to the following optimization

\vspace{-0.2cm}\footnotesize
\begin{eqnarray}
\min_{\{\W_i,\cc_i\}_{i=1}^{2}}\frac{1}{2} \norm{\Phi-\left(\W_2(\W_1\Phi+\cc_1\1^T)+\cc_2\1^T\right)}^2 \nonumber \\ 
{}&&\hspace{-20em}+\frac{\beta}{2}\left(\norm{\W_1}^2+\norm{\W_2}^2\right) \label{eq:obj_hash}
\end{eqnarray}
\begin{equation}
\textrm{s.t. } \W_1\Phi+\cc_1\1^T \in \{-1,1\}^{L\times m} \label{eq:binary-hash}
\end{equation} 

\normalsize 
By solving (\ref{eq:obj_hash}) under the constraint (\ref{eq:binary-hash}), we find hash parameters which satisfy the properties (ii) and (iii), i.e., they not only ensure the binary outputs of the encoder but also minimize the reconstruction error w.r.t. the fixed aggregated representation $\Phi$.
(\ref{eq:obj_hash}) and (\ref{eq:binary-hash}) have same forms as (\ref{eq:obj_ori}) and (\ref{eq:binary0}). Hence, we solve this optimization with the proposed Relaxed Binary Autoencoder (Section \ref{sec:RBA}). Specifically, we use the Algorithm \ref{alg1} for solving $(\W_1,\cc_1,\W_2,\cc_2)$ in which $\Phi$ is used as the training data.

The proposed simultaneous feature aggregating and hashing is presented in the Algorithm \ref{alg2}. In the Algorithm~\ref{alg2}, $\Phi^{(t)}$, $\W_1^{(t)}, \cc_1^{(t)}, \W_2^{(t)}, \cc_2^{(t)}$ are values at $t^{th}$ iteration. After learning $\W_1, \cc_1, \W_2, \cc_2$, given set of local features of a new image, we first compute its aggregated representation $\varphi$ using (\ref{eq:sol_pool}). We then pass $\varphi$ to the encoder %i.e., $h = \W_1\varphi+\cc_1$, 
to compute the binary code.

\begin{algorithm}[!t]
\footnotesize
\caption{Simultaneous feature Aggregating and Hashing (SAH)}
\begin{algorithmic}[1] 
\Require 
	\Statex $\Vs=\{\V_i\}_{i=1}^{m}$: training data; $L$: code length; $T, T_1$: maximum iteration numbers for SAH and RBA (Algorithm \ref{alg1}), respectively; parameters $\lambda, \beta, \gamma, \mu$. 
\Ensure 
	\Statex 
	Parameters $\W_1, \cc_1, \W_2, \cc_2$
	\Statex 
	\State Initialize $\Phi^{(0)}=\{\varphi_i\}_{i=1}^{m}$ with Generalized Max Pooling (\ref{eq:gmp-solution})
	\For{$t = 1 \to T$}
		\State Fix $\Phi^{(t-1)}$, solve $(\W_1^{(t)}, \cc_1^{(t)}, \W_2^{(t)}, \cc_2^{(t)})$ using Algorithm \ref{alg1} (which uses $\Phi^{(t-1)}$ as inputs for training)%\textbf{$(\W,\cc)$-step} (i.e. Algorithm \ref{alg1}).
		\State Fix $(\W_1^{(t)}, \cc_1^{(t)}, \W_2^{(t)}, \cc_2^{(t)})$, solve $\Phi^{(t)}$ using \textbf{$\Phi$-step}.
	\EndFor
	\State Return %$\B_{(max\_iter)}$ and 
	$\W_1^{(T)}, \W_2^{(T)}, \cc_1^{(T)}, \cc_2^{(T)}$
\end{algorithmic}
\label{alg2}
\end{algorithm}

\section{Evaluation of Simultaneous Feature Aggregating and Hashing (SAH)}
\label{sec:evaSAH}
This section evaluates and compares the proposed SAH to the following state-of-the-art unsupervised hashing methods:  
 Iterative Quantization (ITQ)~\cite{DBLP:conf/cvpr/GongL11}, Binary Autoencoder (BA)~\cite{BA_CVPR15} and the proposed RBA, Spherical Hashing (SPH)~\cite{CVPR12:SphericalHashing}, K-means Hashing (KMH)~\cite{DBLP:conf/cvpr/HeWS13}. For all compared methods, we use the implementations and the suggested parameters provided by the authors. The values of $\lambda$, $\beta$, $\gamma$, and $\mu$ are set by cross validation as $10^{-2}$, $10^{-1}$, $10$, and $10^2$, respectively.

\subsection{Dataset}
We conduct experiments on Holidays \cite{herve_ijcv2010} and Oxford5k \cite{Philbin07-cvpr-2007} datasets which are widely used in evaluating image retrieval systems \cite{DBLP:conf/cvpr/JegouZ14,DBLP:conf/cvpr/ArandjelovicZ13,herve_cvpr2010}.

\textbf{Holidays:} The Holidays dataset  \cite{herve_ijcv2010} consists of 1,491 images of different locations and objects, 500 of them being used as queries. Follow standard configuration~\cite{DBLP:conf/cvpr/JegouZ14,DBLP:conf/cvpr/ArandjelovicZ13}, when evaluating, we remove the query from the ranked list. For the training dataset, we follow \cite{DBLP:conf/cvpr/JegouZ14,DBLP:conf/cvpr/ArandjelovicZ13}, i.e., using 10k images from the independent dataset Flickr60k provided with the Holidays.

\textbf{Holidays+Flickr100k:} In order to evaluate the proposed method on large scale, we merge Holidays dataset with 100k images downloaded from Flickr \cite{herve_eccv2008}, forming the Holidays+Flickr100k dataset. This dataset uses the same training dataset with Holidays. 

\textbf{Oxford5k:} The Oxford5k dataset \cite{Philbin07-cvpr-2007} consists of 5,063 images of buildings and 55 query images corresponding to 11 distinct buildings in Oxford. We follow standard protocol \cite{DBLP:conf/cvpr/JegouZ14,DBLP:conf/cvpr/ArandjelovicZ13}: the bounding boxes of the region of interest are cropped and then used as the queries. As standardly done in the literature, for the learning, we use the Paris6k dataset \cite{Philbin08_cvpr}.
%\vspace{-0.1cm}

The ground truth of queries have been provided with the datasets \cite{herve_ijcv2010,Philbin07-cvpr-2007}. Follow the state of the art \cite{DBLP:conf/cvpr/GongL11,BA_CVPR15}, we evaluate the performance of methods with mAP. 
%\subsection{Implementation details}
\subsection{Experiments with SIFT features}
%For the local image representation, 
Follow state-of-the-art image retrieval systems \cite{DBLP:conf/cvpr/JegouZ14,herve_cvpr2010,F-FAemb}, to describe images, we extract SIFT local descriptors \cite{SIFT_Lowe} on Hessian-affine regions \cite{mikolajczyk_scale_2004}. RootSIFT variant \cite{DBLP:conf/cvpr/ArandjelovicZ12} is used in all our experiments. 
 Furthermore, instead of directly using SIFT local features,
%which are lack of discriminability \cite{DBLP:conf/cvpr/JegouZ14,DBLP:conf/iccv/BabenkoL15}, 
%man
as a common practice, 
we enhance their discriminative power by embedding them into high dimensional space (i.e., 1024 dimensions) with the state-of-the-art triangulation embedding~\cite{DBLP:conf/cvpr/JegouZ14}.  As results, the set of triangulation embedded vectors $\Vs=\{\V_i\}_{i=1}^m$ is used as the input for the proposed SAH. 
In order to make a fair comparison to other methods, we aggregate the triangulation embedded vectors  with GMP \cite{gmp} and use the resulted vectors %resulting $\varphi$ and use this $\varphi$ representation 
as inputs for compared hashing methods.

\begin{figure}[!t]
\centering
\includegraphics[height=4cm, width=6cm]{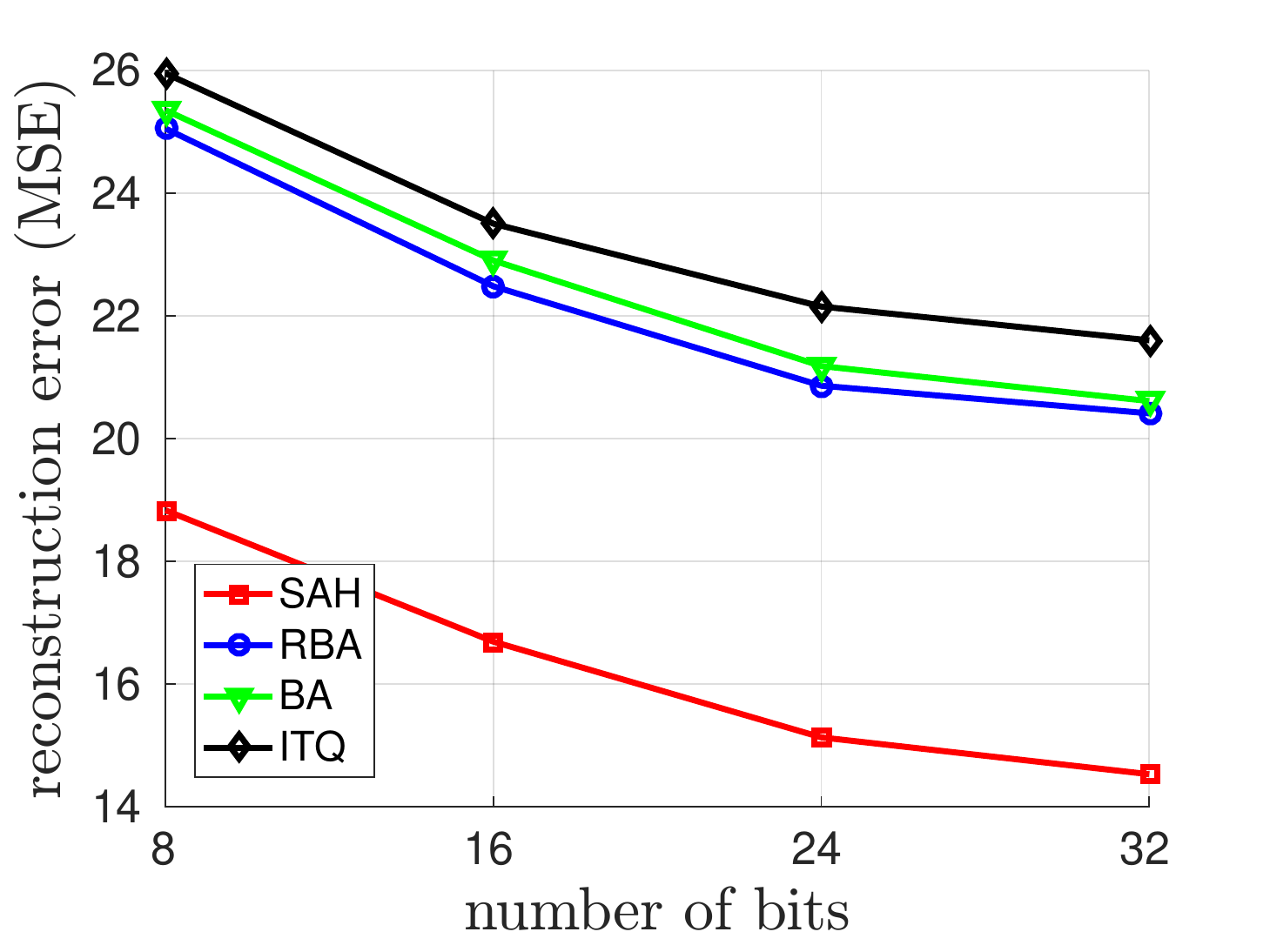}
\caption[]{Reconstruction error (Mean Square Error -- MSE) comparison of different methods on Oxford5k dataset}
\label{fig:res_oxford5k}
\end{figure}
\subsubsection{Reconstruction comparison}
In this experiment, we evaluate the reconstruction capacity of binary codes produced by different methods: ITQ \cite{DBLP:conf/cvpr/GongL11}, BA \cite{BA_CVPR15}, RBA,  and SAH. We compute the average reconstruction error on the Oxford5k dataset.

For ITQ, BA, and RBA, given the binary codes $\Z$ of the testing data (Oxford5k), the reconstructed testing data is computed by $\X_{res} = \W_2 \Z + \cc_2 \1^T$, where ($\W_2, \cc_2$) is  decoder. Note that the decoder is available in the design of BA/RBA and is learned in learning process. For ITQ, there is no decoder in its design, hence we follow \cite{BA_CVPR15}, i.e., we compute the optimal linear decoder ($\W_2, \cc_2$) using the binary codes of the training data (Paris6k). 

For SAH, given the binary codes $\Z$, we use the learned encoder and decoder to compute the aggregated representations $\Phi$ by using (\ref{eq:sol_pool}). The reconstruction of $\Phi$ is computed by using the decoder as $\Phi_{res} = \W_2 \Z + \cc_2 \1^T$.

Figure \ref{fig:res_oxford5k} shows that BA and RBA are comparable while SAH dominates all other methods in term of reconstruction error. This confirms the benefit of the jointly learning of aggregating and hashing in the proposed SAH. 

\begin{figure*}[!t]
\centering

\subfigure[Holidays]{
       \includegraphics[scale=0.33]{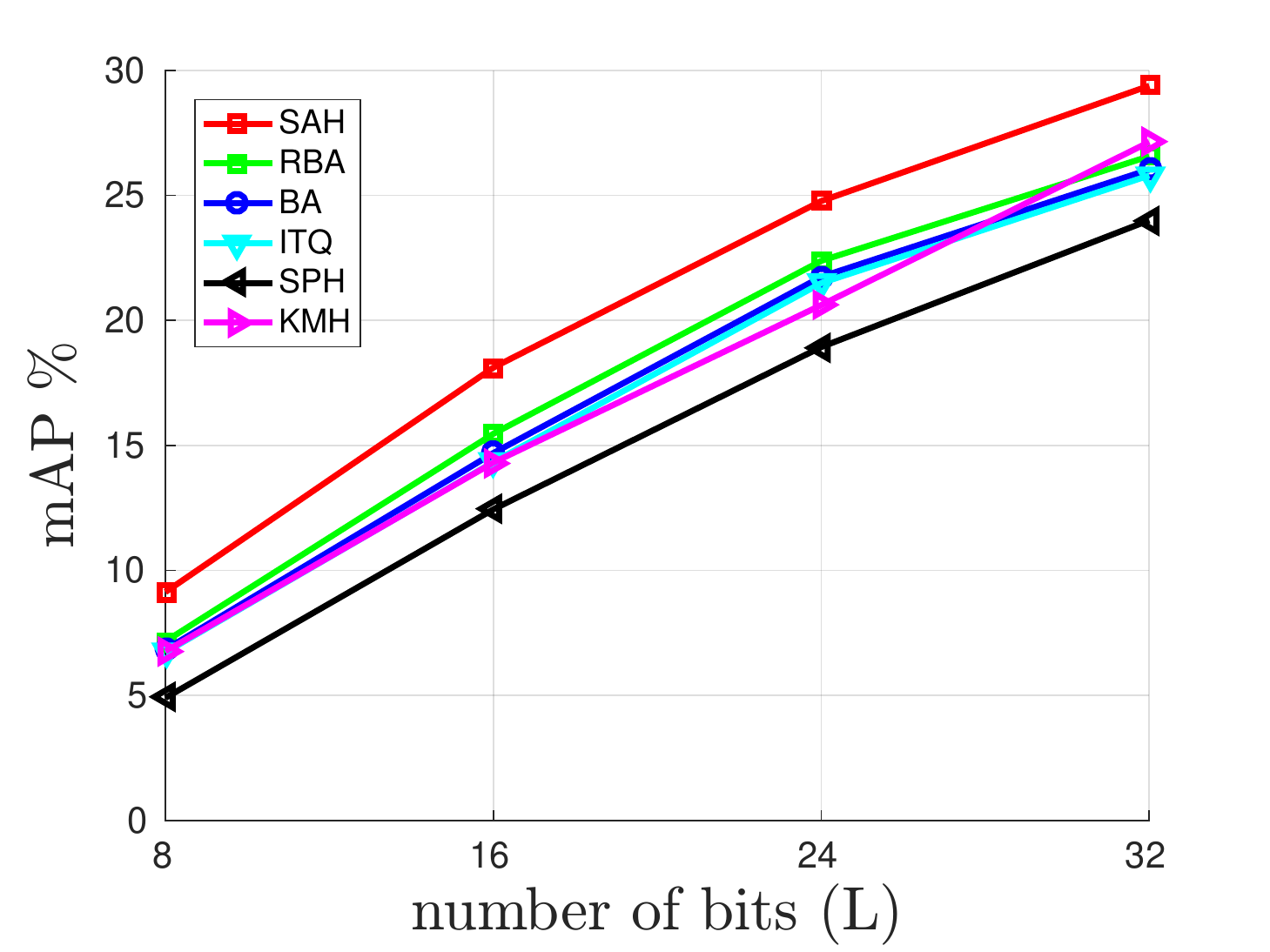}
       \label{fig:holidays_mAP_SIFT}
}
\subfigure[Oxford5k]{
       \includegraphics[scale=0.33]{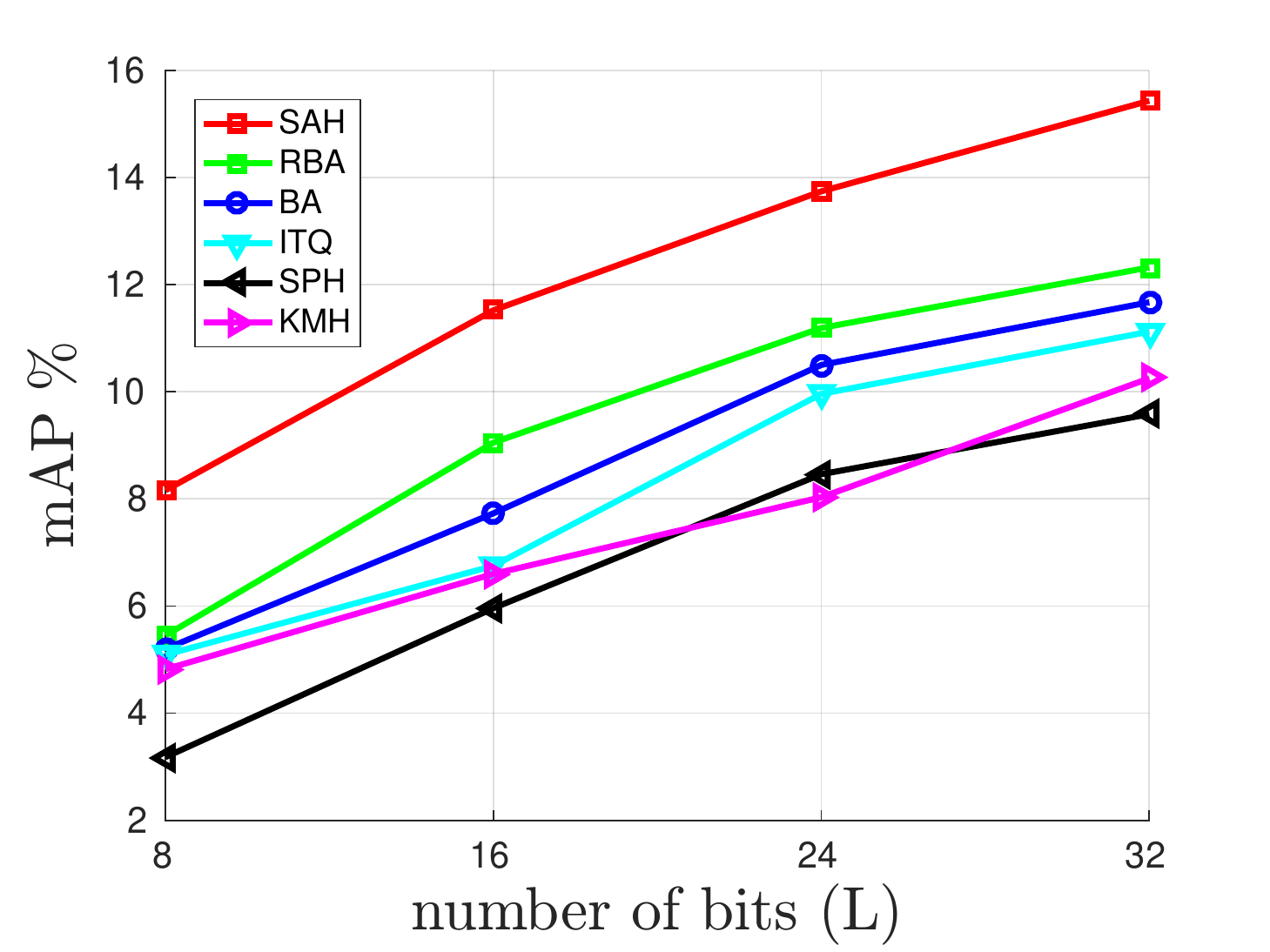} 
       \label{fig:oxford5k_mAP_SIFT}
}
\subfigure[Holidays+Flickr100k]{ 
       \includegraphics[scale=0.33]{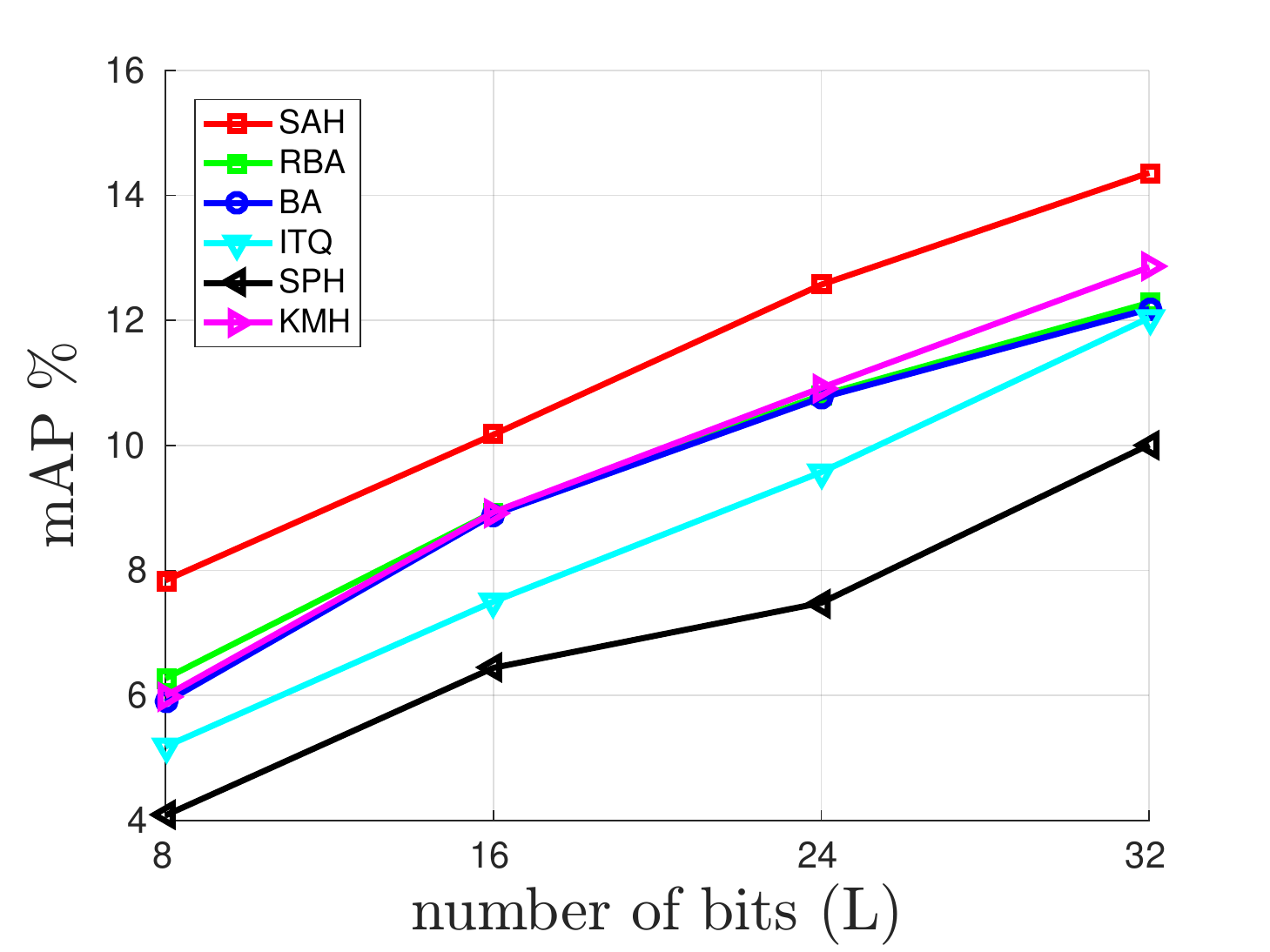} 
       \label{fig:holidays100k_mAP_SIFT}
}
\caption[]{mAP (\%) comparison between SAH and state-of-the-art unsupervised hashing methods when using SIFT features on Holidays, Oxford5k, and Holidays+Flickr100k. GMP \cite{gmp} is used to aggregate embedded local SIFT vectors to produce global vectors used as inputs for the compared methods.}
\label{fig:lph-soa}
\end{figure*}

\subsubsection{Retrieval results}

Figure \ref{fig:lph-soa} shows the comparative mAP between compared methods when using SIFT features. We find the following observations are consistent on three datasets. The proposed RBA is competitive or slightly outperforms BA \cite{BA_CVPR15}, especially on Oxford5k dataset. The proposed  SAH improves other methods by  a fair margin. The improvement is more clear on Holidays and Oxford5k, e.g., SAH outperforms the most competitor RBA 2\%-3\% mAP at all code lengths. 

%\vspace{-0.1cm}
\subsection{Experiments with CNN feature maps}
Recently, in \cite{DBLP:journals/corr/ToliasSJ15,DBLP:conf/iccv/BabenkoL15,DBLP:journals/corr/AzizpourRSMC14} the authors showed that the activations from the convolutional layers of a convolutional neural network (CNN) can be interpreted as local features describing image regions. 
  Motivated by those works, in this section we perform the experiments in which  activations of a convolutional layer from a pre-trained CNN are used as an alternative to SIFT features. It is worth noting that our work is the first one that evaluates hashing on the image representation aggregated from convolutional features. 
\subsubsection{Evaluation protocol}
\label{subsub:cnnfea}
\red{
We extract the activations of the $5^{th}$ convolutional layer (the last convolutional layer) of the pre-trained VGG-16 network \cite{Simonyan14c}. Given an image, the activations form a 3D tensor of $W\times H \times C$, where $C=512$ which is number of feature maps and $W=H=37$ which is spatial size of the last convolutional layer. By using this setting, we consider that each image is represented by $37\times37=1,369$ local feature vectors with dimensionality $512$.}
In \cite{DBLP:conf/iccv/BabenkoL15}, the authors showed that the convolutional features are discriminative,  hence the embedding step is not needed for these features. Therefore, we directly use the convolutional features as the inputs for the proposed SAH. In order to make a fair comparison between SAH and other hashing methods, we aggregate the convolutional features with GMP \cite{gmp} and use the resulted vectors as the inputs for compared hashing methods. 

\begin{figure*}[!t]
\centering
\vspace{-0.7cm}
\subfigure[Holidays]{
       \includegraphics[scale=0.33]{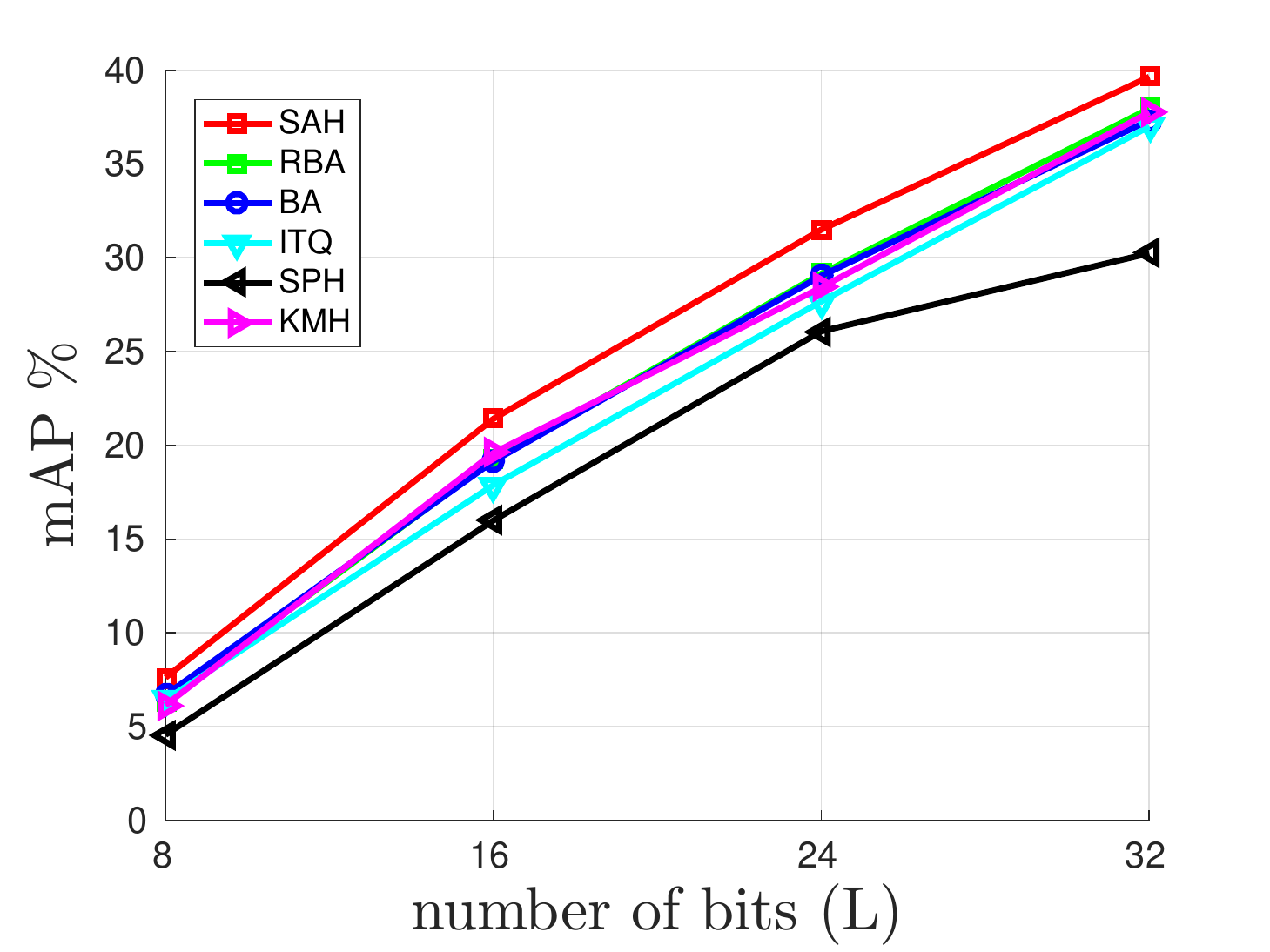}
       \label{fig:holidays_mAP_conv}
}
\subfigure[Oxford5k]{
       \includegraphics[scale=0.33]{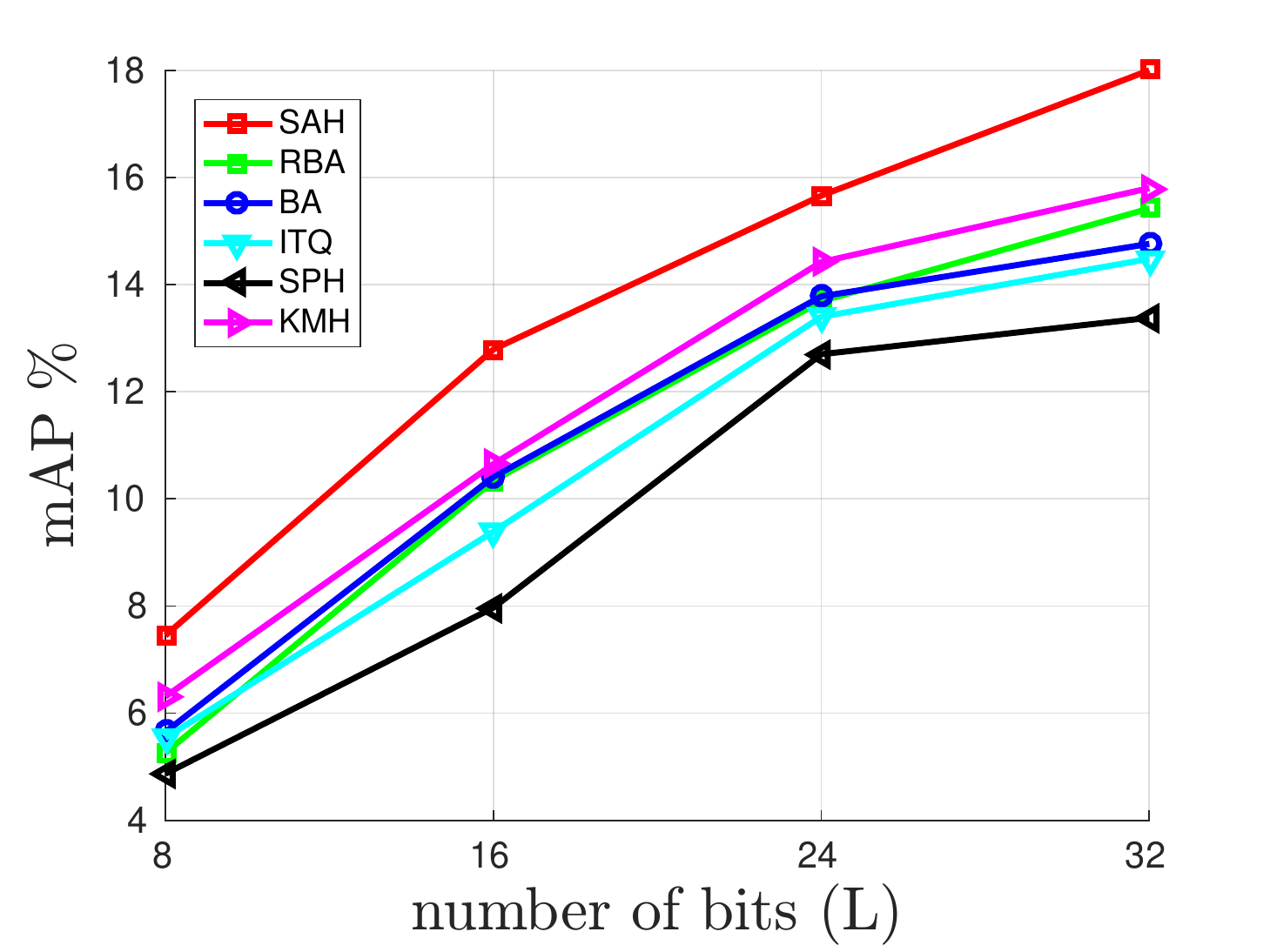} 
       \label{fig:oxford5k_mAP_conv}
}
\subfigure[Holidays+Flickr100k]{ 
       \includegraphics[scale=0.33]{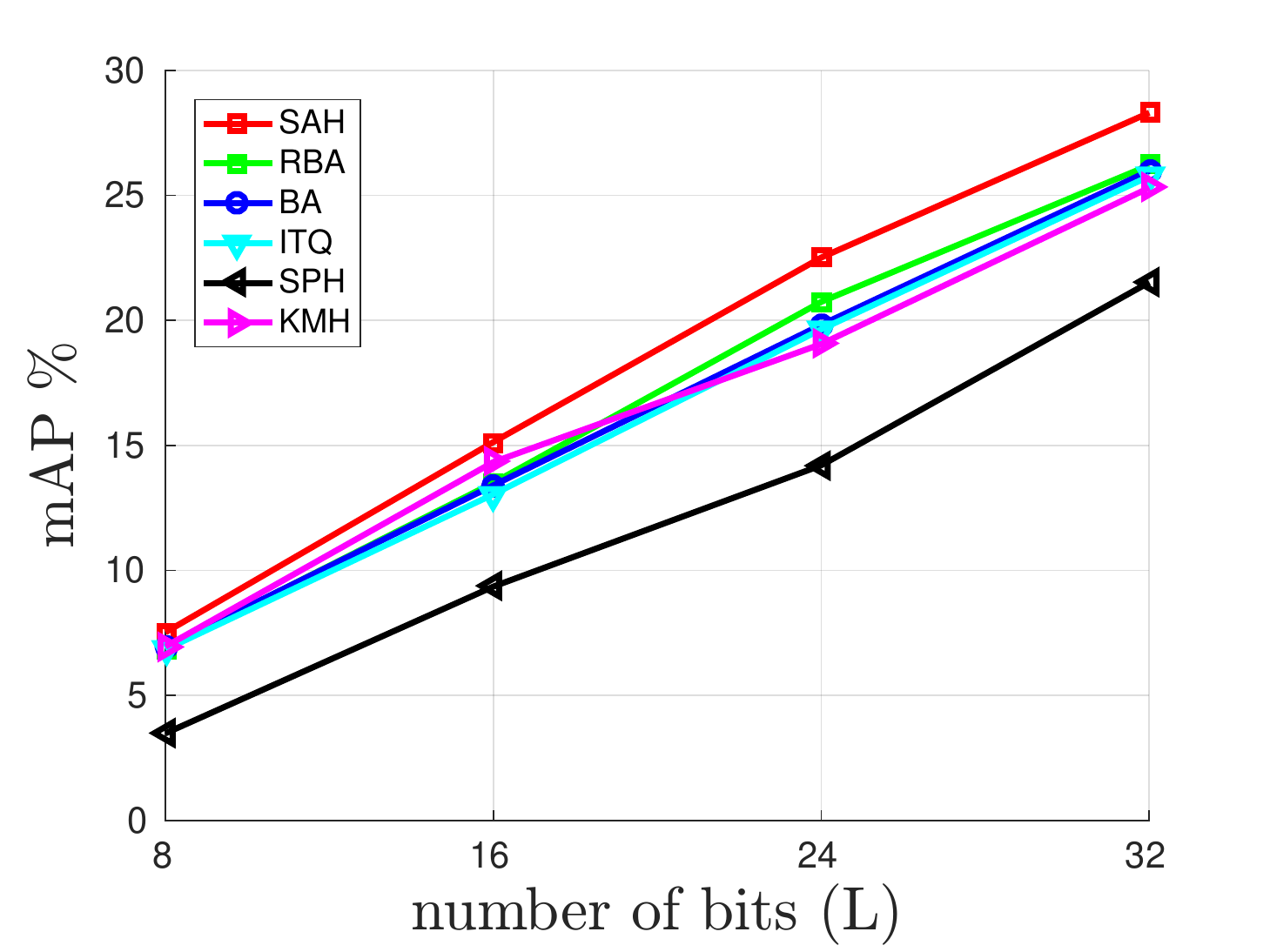} 
       \label{fig:holidays100k_mAP_conv}
}
\caption[]{mAP (\%) comparison between SAH and state-of-the-art unsupervised hashing methods when using convolutional features on Holidays, Oxford5k, and Holidays+Flickr100k. Note that GMP \cite{gmp} is used to aggregate local convolutional feature vectors to produce global vectors as inputs for the compared methods.}
\label{fig:lph-soa_CNN}
\end{figure*}
% \vspace{-0.5em}
\begin{figure*}[!t]
\centering
%\vspace{-0.1cm}
\subfigure[Holidays]{
       \includegraphics[scale=0.33]{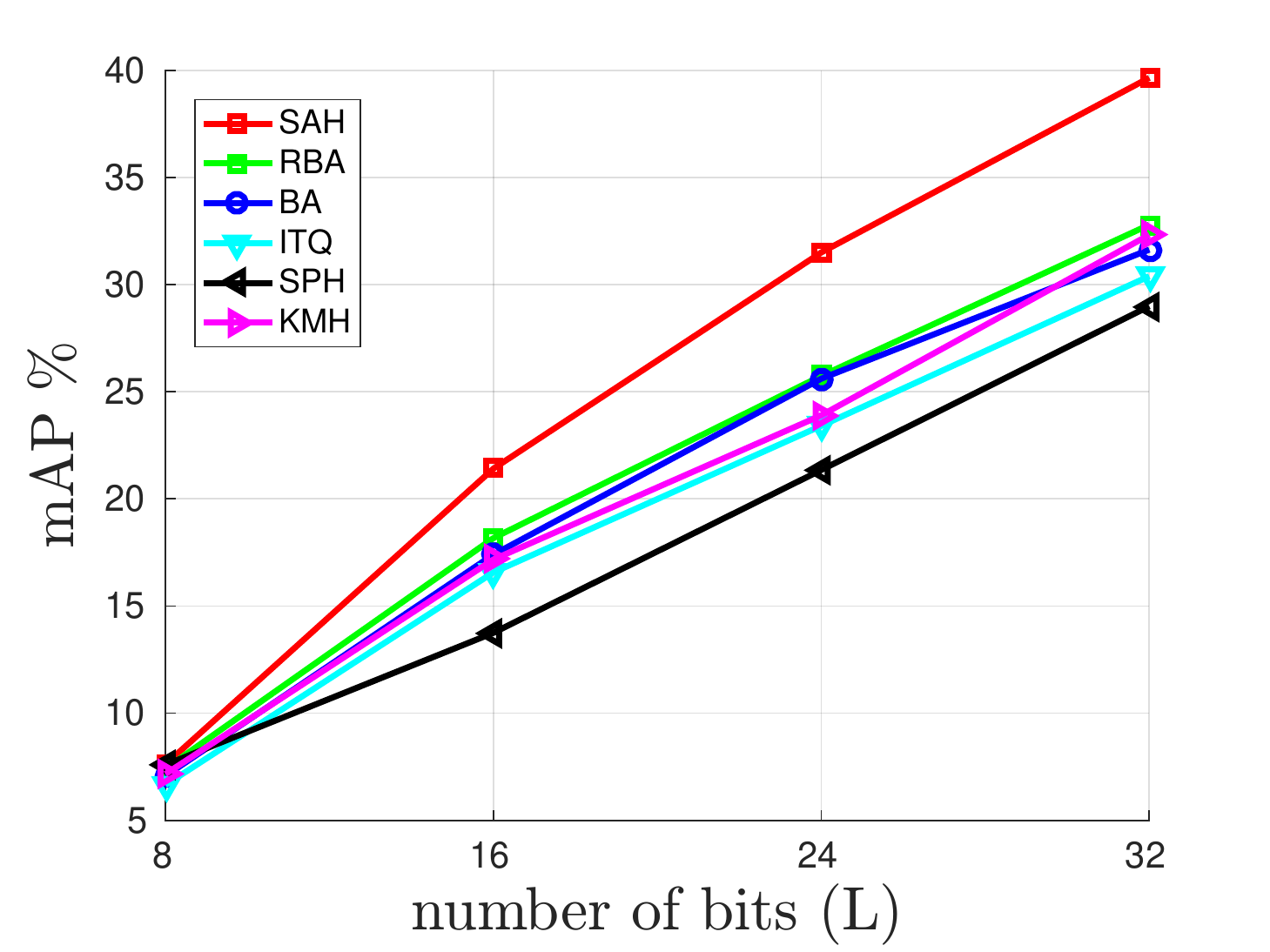}
       \label{fig:holidays_mAP_FC7}
}
\subfigure[Oxford5k]{
       \includegraphics[scale=0.33]{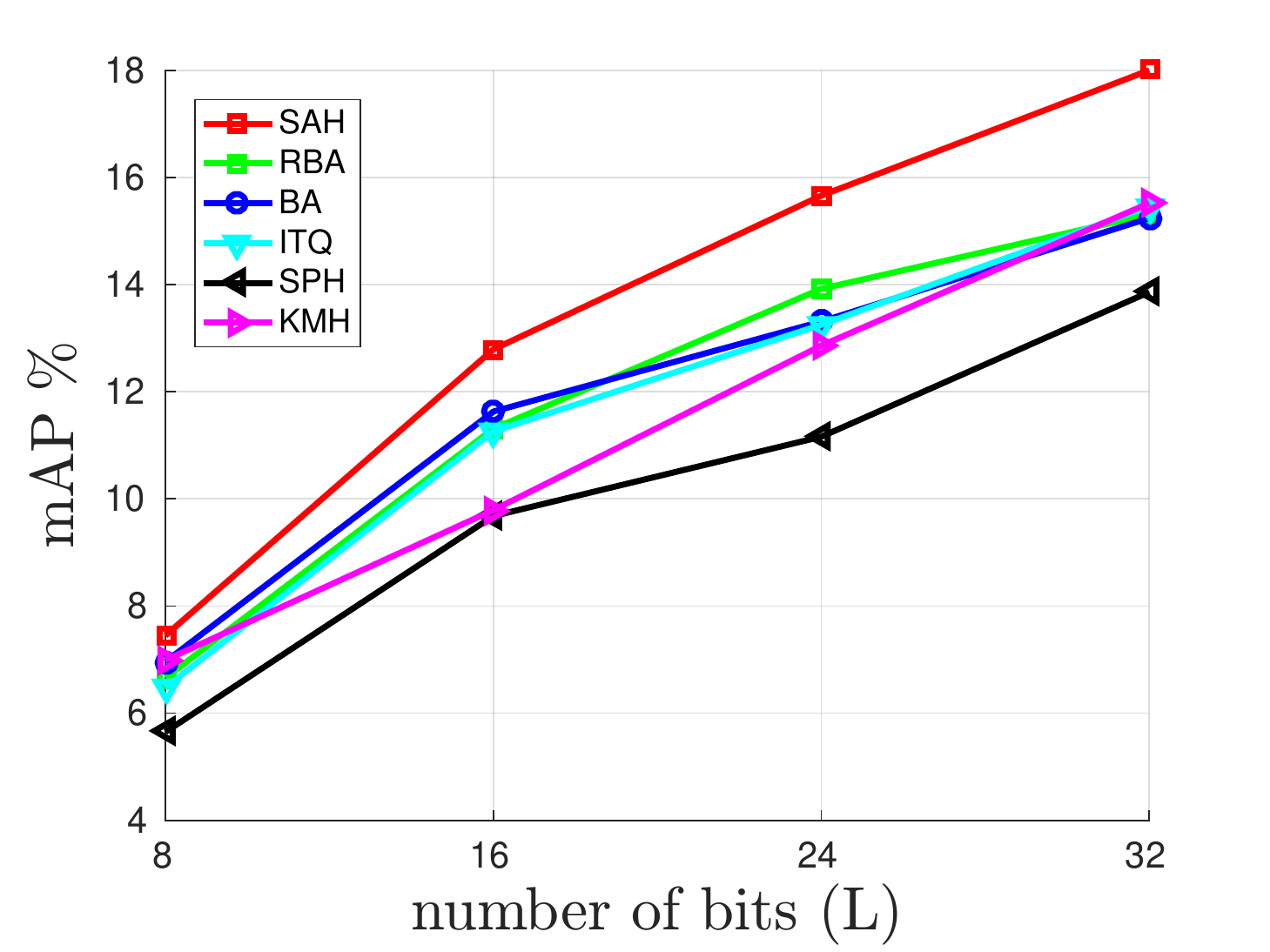} 
       \label{fig:oxford5k_mAP_FC7}
}
\subfigure[Holidays+Flickr100k]{ 
       \includegraphics[scale=0.33]{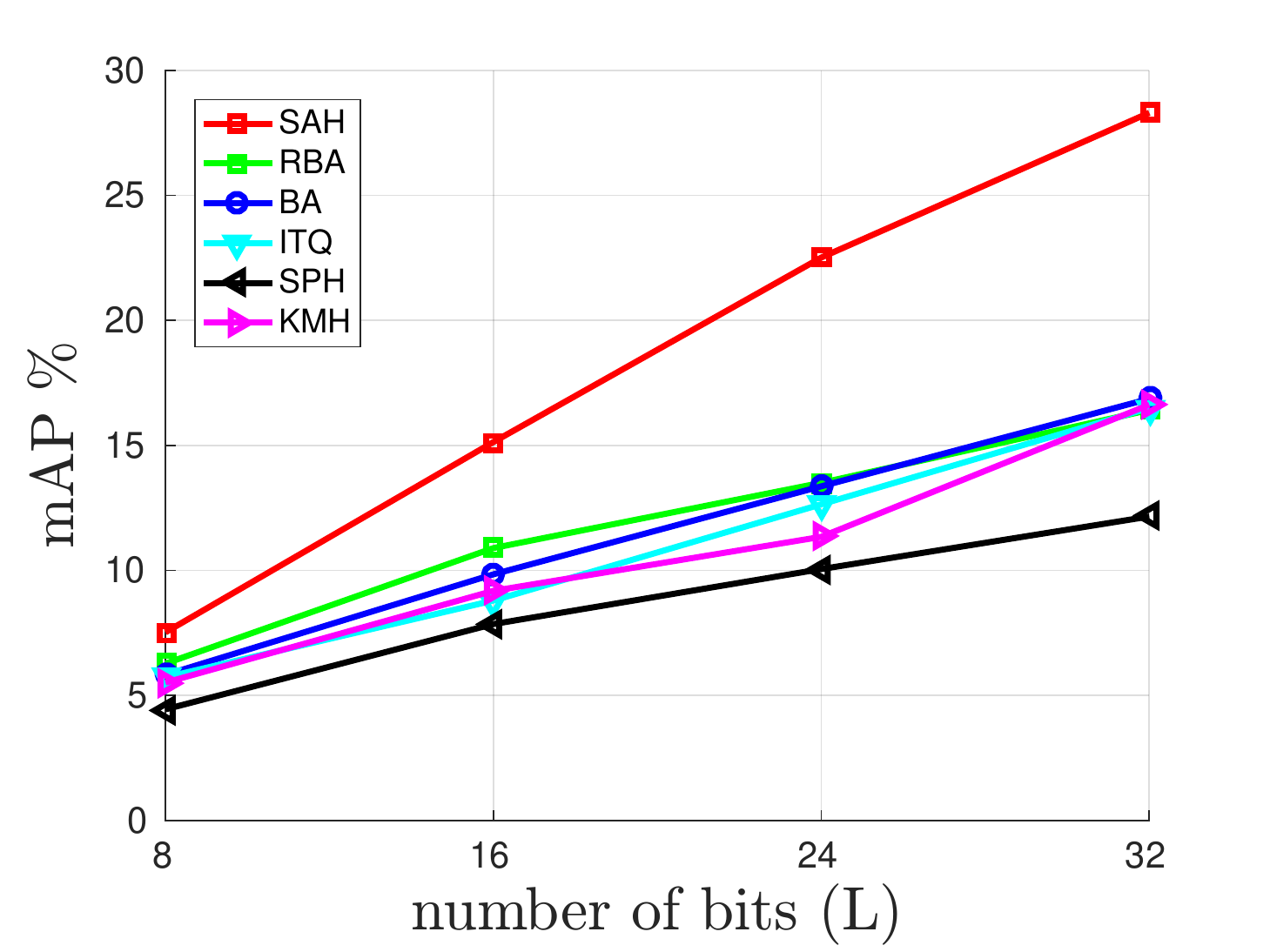} 
       \label{fig:holidays100k_mAP_FC7}
}
\caption[]{mAP (\%) comparison between SAH and state-of-the-art unsupervised hashing methods using fully-connected features on Holidays, Oxford5k, and Holidays+Flickr100k. \red{Note that SAH still takes the convolutional features as the input}.}
\label{fig:lph-soa_CNN_FC7}
\end{figure*}

\subsubsection{Retrieval results}
Figure \ref{fig:lph-soa_CNN} shows the comparative mAP between methods. 
The results show that BA \cite{BA_CVPR15}, KMH \cite{DBLP:conf/cvpr/HeWS13} and RBA achieve comparative results. The results also clearly show that the proposed SAH outperforms other methods by a fair margin. 
The improvements are more clear with longer codes, e.g., SAH outperforms BA \cite{BA_CVPR15} 2\%-3\% mAP at $L = 32$ on three datasets. 
It is worth noting from Figure \ref{fig:lph-soa_CNN} and Figure \ref{fig:lph-soa} that at low code length, i.e., $L=8$, SIFT features and convolutional features give comparable results. However, when increasing the code length, the convolutional features significantly improves over the SIFT features, especially on Holidays and Holidays+Flickr100k datasets. For example, for SAH on Holidays+Flickr100k, the convolutional features improves mAP over the SIFT features about 5\%,  10\%, 14\% for $L=16,24$ and $32$, respectively.

\subsection{Comparison with fully-connected features}
\subsubsection{Evaluation protocol}
In \cite{DBLP:conf/cvpr/RazavianASC14}, the authors showed that for image retrieval problem, using fully-connected features produced by a CNN outperforms most hand-crafted features such as VLAD \cite{herve_cvpr2010}, Fisher \cite{DBLP:conf/cvpr/PerronninD07}.  
In this section, we compare the proposed SAH %which takes the convolutional features as inputs 
with state-of-the-art unsupervised hashing methods which take the fully-connected features (e.g., outputs of the $7^{th}$ fully-connected layer from the pre-trained VGG-16 network \cite{Simonyan14c}) as inputs. 
For the proposed SAH, we take the convolutional features of the $5^{th}$ convolutional layer of the same pre-trained VGG-16 network as inputs to demonstrate the benefit of the jointly learning of aggregating and hashing.

\subsubsection{Retrieval results}
Figure \ref{fig:lph-soa_CNN_FC7} presents the comparative mAP between methods. At low code length, i.e., $L=8$, SAH is competitive to other methods. However, when increasing the code length, SAH outperforms compared methods a large margin. The significant improvements are shown on Holidays and Holidays+Flickr100k datasets, e.g., at $L=32$, the improvements of SAH over BA \cite{BA_CVPR15} are 8\% and 11.4\% on Holidays and Holidays+Flickr100k, respectively. 

\smallskip
\red{
From Figures \ref{fig:lph-soa_CNN} and \ref{fig:lph-soa_CNN_FC7}, we can observe that for the compared unsupervised hashing methods, using the aggregated local convolutional features instead of fully-connected features can help to achieve significant gains in performance. This indicates the clear advantage of using convolutional features for the retrieval task. 
And hence, jointly learning to aggregate convolutional features and hashing is very beneficial.}
\red{
Additionally, we summarize the retrieval performance of SAH when using SIFT and convolutional features (Conv.) on different datasets in Table \ref{tab:SAH_SIFT_conv}. Generally, the SAH with convolutional features unsurprisingly achieves better performance than SAH with SIFT.  Furthermore, the performance gaps increase as code lengths increase.
}

\begin{table}[!t]
\centering
\footnotesize
%\small
\caption{SAH retrieval performance (mAP \%) when using SIFT and convolutional features as inputs on Holidays, Oxford5k, and Holidays+Flickr100k datasets.} 
\begin{tabular}{|l|c|c|c|c|c|} 
\hline
Dataset & Feature & 8 bits & 16 bits & 24 bits & 32 bits\\ \hline
\multirow{2}{*}{Holidays} & SIFT & 9.13 & 18.11 & 24.77 & 29.38\\ \cline{2-6}
 & Conv. & 7.61 & 21.41 & 31.50 & 39.65\\ \hline
\multirow{2}{*}{Oxford5k} & SIFT & 8.16 & 11.52 & 13.74 & 15.43\\ \cline{2-6}
& Conv. & 7.45 & 12.79 & 15.66 & 18.00\\ \hline
Holidays & SIFT & 7.84 & 10.17 & 12.57 & 14.36\\ \cline{2-6}
+Flickr100k & Conv. & 7.52 & 15.13 & 22.51 & 28.30\\ \hline
\end{tabular}
\label{tab:SAH_SIFT_conv}
\end{table}

\subsection{Comparison with the recent deep learning-based unsupervised hashing methods}
\subsubsection{DeepBit \cite{deepbit2016,deepbit2018}}
%\vspace{0.3cm}
%\textbf{Comparison with DeepBit \cite{deepbit2016}}
In \cite{deepbit2016}, the authors  proposed an end-to-end CNN-based unsupervised hashing approach, named DeepBit. To the best of our knowledge, this is the first work using end-to-end CNN for unsupervised hashing. Starting with the pre-trained VGG network \cite{Simonyan14c}, the authors replaced the softmax layer of VGG with their binary layer and enforced several criteria on the binary codes learned at the binary layer, i.e., binary codes should: minimize the quantization loss with the output of the last VGG's fully connected layer, be distributed evenly, be invariant to rotation. 
Their network is fine-tuned using 50k training samples of CIFAR10. Note that as DeepBit is unsupervised, no label information is used during fine-tuning. 
Recently, in~\cite{deepbit2018}, the authors have improved their DeepBit by using data augmentation to enhance the scale invariance and translation invariance of learned binary codes. 
The comparative mAP between DeepBit, its improved version (DeepBit-imp.) and other methods on the top $1,000$ returned images (with the class label ground truth) on the testing set of CIFAR10 is cited in the top part of Table \ref{tab:deepbit}. 

\red{
\subsubsection{GraphBit \cite{graphbit}}
In \cite{graphbit}, the authors proposed to simultaneously learn deep binary descriptors and the structure of a graph, called GraphBit, which represents the interactions (as edges) among different bits. In specific, each bit of the binary descriptor is learned to maximize its mutual information with the input features, while the GraphBit provides additional information where each bit chooses to be instructed by either only inputs or with additional related bits. 
The retrieval performance of GraphBit in mAP for $1,000$ returned images (as similar to DeepBit) is also presented in the top part of Table \ref{tab:deepbit}. Note that GraphBit also uses the pretrained VGG \cite{Simonyan14c} as the initial model.
}

\smallskip
In DeepBit~\cite{deepbit2016} and GraphBit \cite{graphbit}, the authors reported results of ITQ, KMH, SPH when the GIST features are used.  
Here, we also evaluate those three hashing methods on the features extracted from the activations of the last fully connected layer of the same pre-trained VGG \cite{Simonyan14c} with the same setting. These results, i.e., ITQ-CNN, KMH-CNN, SPH-CNN, are presented in the bottom part of Table \ref{tab:deepbit}. It clearly shows that using fully-connected instead of GIST, ITQ-CNN, KMH-CNN, SPH-CNN have improvements and outperforms DeepBit and GraphBit.

In order to evaluate the proposed SAH, we extract the activations of the last convolutional layer of the same pre-trained VGG and use them as inputs. Similar to DeepBit, we report the mAP on the top $1,000$ returned images.
% for SAH. 
The results of SAH presented in the last row in Table \ref{tab:deepbit} show that at the same code length, SAH significantly outperforms the recent end-to-end works  DeepBit\cite{deepbit2016}, DeepBit  improved version~\cite{deepbit2018}, and GraphBit~\cite{graphbit}. 
Furthermore, SAH also outperforms ITQ-CNN, KMH-CNN, SPH-CNN with  fair margins. 

\begin{table}[!t]
\centering
\footnotesize
%\small
\caption{Comparison between SAH with DeepBit \cite{deepbit2016}, DeepBit improved version~\cite{deepbit2018}, GraphBit~\cite{graphbit}, and other unsupervised hashing methods on CIFAR10.   The mAP (\%) is computed on the top 1000 returned images.} 
\begin{tabular}{|l|c|c|c|c|c|} 
\hline
Method    &16 bits &32 bits &64 bits\\ \hline
ITQ \cite{DBLP:conf/cvpr/GongL11} &15.67 &16.20  &16.64\\ \hline
KMH \cite{DBLP:conf/cvpr/HeWS13} &13.59 &13.93  &14.46\\ \hline
SPH \cite{CVPR12:SphericalHashing} &13.98 &14.58  &15.38\\ \hline
DeepBit \cite{deepbit2016} &19.43 &24.86 &27.73\\ \hline
DeepBit-imp.\cite{deepbit2018} &26.36 &27.92 &34.05 \\ \hline
GraphBit \cite{graphbit} & 32.15 & 36.74 & 39.90 \\ \hline
 \Xhline{3\arrayrulewidth}
ITQ-CNN &38.52	&41.39 &44.17\\ \hline
KMH-CNN &36.02	&38.18 &40.11\\ \hline
SPH-CNN &30.19	&35.63 &39.23\\ \hline
SAH &\textbf{41.75} &\textbf{45.56} &\textbf{47.36}\\ \hline
\end{tabular}
\label{tab:deepbit}
\end{table}

%\vspace{-0.5cm}
\subsubsection{Similarity-Adaptive Deep Hashing~\cite{SADH}}
\begin{table}[!t]
\centering
\footnotesize
%\small
\caption{Comparison between SAH with SADH \cite{SADH} on CIFAR10. The mAP (\%) is computed on all returned images. }  
\begin{tabular}{|l|c|c|c|c|c|} 
\hline
Method    &16 bits &32 bits &64 bits\\ \hline
SADH~\cite{SADH} &38.70	&38.49 &37.68\\ \hline
SAH  &36.76 &37.85 &38.52\\ \hline
\end{tabular}
\label{tab:SADH}
%\vspace{-0.5cm}
\end{table}
Recently, in~\cite{SADH} the authors proposed Similarity-Adaptive Deep Hashing (SADH), an unsupervised hashing method which is based on deep learning. SADH alternatively proceeds over three training modules: deep hash model training, similarity graph updating and binary code optimization. The first module, which is a deep hash model, has a Euclidean loss layer, which measures the discrepancy between the outputs of the deep model and the binary codes learned by the third module. The similarity graph updating module updates the similarity matrix between training images using the current learned deep features of the first module. The binary code optimization module learns binary codes using a graph-based approach, in which the similarity matrix outputted by the second module is used as input. Table~\ref{tab:SADH} presents comparative results between the proposed SAH and SADH~\cite{SADH} on the CIFAR10 dataset. The experiment setting is same as SADH~\cite{SADH}, in which 100 images per class is randomly sampled and is used as the query set. The rest images are used as the training set. The mAP is computed on all returned images. 

\red{The results in Table \ref{tab:SADH} show that SAH achieves competitive results with SADH. 
 Although SADH slightly outperforms SAH at low code length (e.g., 16 bits), at larger code lengths, both methods
achieve  comparable  mAP.  However,  it  is  worth  noting  that  performance  of  SADH  is  saturated at  16  bits. Its  performance  is  decreased  when  increasing  code  length.  This  fact  indicates  that SADH potentially suffers from the overfitting problem as more bits are used. Contrary to SADH, the proposed SAH gets improvements when increasing the code length and slightly outperforms
SADH at 64 bits. It means that the proposed method is not overfitted over the training set and it
is well generalized. 
}

\subsection{Complexity analysis}
\red{

At testing stage, to compute the binary codes, the proposed SAH  includes two steps: (i) compute the aggregating (global) representation $\varphi$ (Eq. \ref{eq:sol_pool}) and (ii) encode $\varphi$ to compute the binary code. 
In specific, the (asymptotic) complexity for computing ( \ref{eq:sol_pool}) is $\mathcal{O} (\max(D^3,D^2n_i))$, where $D$ is dimension of the local features and $n_i$ is the number of local features in the image. This  complexity is similar to the  complexity for computing the original Generalized Max Pooling (GMP) (Eq. \ref{eq:gmp-solution}). Additionally, the  complexity of the encoding step is $\mathcal{O}(DL)$ which is much smaller than the complexity to compute the aggregated features. We note that given a set of local features, the aggregated features are computed using Eq. (\ref{eq:gmp-solution}) before fed to other hashing methods, e.g., ITQ, BA, SPH. That means that the  complexity of SAH and other methods are similar when computing the global representation. 
However, we observe that the running time to compute GMP (Eq. \ref{eq:gmp-solution}) is a bit faster than the one of computing $\varphi$ (Eq.  \ref{eq:sol_pool}). Table \ref{tab:computation_time} shows the computational time (ms) of different steps of SAH and other hashing methods. 
We can observe that SAH takes longer to compute $\varphi$ than GMP. However, it is still very fast and practical for large scale retrieval systems. In addition, Table \ref{tab:computation_time} shows that the most expensive step is to extract the convolutional features.

Regarding the space requirements, SAH requires to store both encoder and decoder matrices, i.e., $2\times D\times L$ floating-point numbers where $D$ is the dimension of local features and $L$ is the code length. For local convolutional features (i.e., section V.C), $D=512$.  The $2\times D\times L$ floating-point space requirement is bigger than the $D\times L$ floating-point  space requirement of ITQ, BA. Nevertheless, these space requirements are very small and negligible in comparison with the storage requirement for the pretrained CNN model and the representation of input data.

\begin{table}[!t]
\centering
\footnotesize
%\small
\caption{The computational time (ms) of different steps of SAH and other hashing methods. The convolutional feature extraction is conducted with a TITAN X GPU using MatConvNet~\cite{vedaldi15matconvnet}.}

\begin{tabular}{|l|c|c|c|c|c|} 
\hline
Method    & SAH & Other methods \\ \hline
Extract conv. features & \multicolumn{2}{c|}{$\approx 80$ ms} \\ \hline
Compute $\varphi$ & 6.3 ms (Eq. \ref{eq:sol_pool}) & 4.1 ms (Eq. \ref{eq:gmp-solution})\\ \hline
Compute bin. codes from $\varphi$ & $<1$ ms & $<1$ ms\\ \hline
\end{tabular}
\label{tab:computation_time}
\end{table}
}

\section{Simultaneous Feature Aggregating and Supervised Hashing (SASH)}
\label{sec:SASH}
An advantage of the formulation of SAH (\ref{eq:obj_ori_join}) is the flexibility. When the data label is available, it is possible to extend the unsupervised SAH into its supervised version. In this section, we present the supervised version of SAH, namely simultaneous feature aggregating and supervised hashing (SASH).
\subsection{Formulation}
In order to take advantage of the label information, 
instead of learning binary codes which provide a good reconstruction of aggregated features as in unsupervised version (Section~\ref{sec:SAH}), here we aim to learn binary codes which provide a good reconstruction of label vectors. Specifically, we propose to minimize the following constrained objective function

\vspace{-0.3cm}\footnotesize
\begin{eqnarray}
{}&&\hspace{-3em}\min_{\W_1,\cc_1,\W_2,\cc_2,\Phi} \frac{1}{2} \norm{\Y-\left(\W_2(\W_1\Phi+\cc_1\1^T)+\cc_2\1^T\right)}^2 \nonumber \\ 
{}&&\hspace{-4em}+\frac{\beta}{2}\left(\norm{\W_1}^2+\norm{\W_2}^2\right)+\frac{\gamma}{2}\sum_{i=1}^{m}\left(\norm{\V_i^T\varphi_i-\1}^2+\mu\norm{\varphi_i}^2 \right) \label{eq:obj_ori_join_sup}
\end{eqnarray}
\begin{equation}
\textrm{s.t. } \W_1\Phi+\cc_1\1^T \in \{-1,1\}^{L\times m} \label{eq:binary_join_sup}
\end{equation} 
\normalsize 
 where $\Y = \{\y_i\}_{i=1}^m \in \R^{C \times m}$; $C$ is the number of classes; $\y_i \in \R^{C\times 1}$ is the label vector of sample $i$, in which the index of the maximum element indicates the class of the sample.   
The proposed constrained objective function (\ref{eq:obj_ori_join_sup}) has a clear meaning. The term $\W_1\Phi+\cc_1\1^T$ can be seen as a feature mapping (or encoder) which maps learned aggregated features $\Phi$ to binary codes, thanks to the constraint  (\ref{eq:binary_join_sup}). $(\W_2,\cc_2$) can be seen as a linear classifier (or decoder). It takes the codes ($\W_1\Phi+\cc_1\1^T$) as inputs and minimizes the $l_2$ loss w.r.t. the label. The second term of (\ref{eq:obj_ori_join_sup}) is the regularization on the model weights. Finally, the third term of (\ref{eq:obj_ori_join_sup}) encourages the aggregating property, i.e., equaling the similarities between the learned aggregated representation $\varphi_i$ and each column of $\V_i$. 

\subsection{Optimization}
In order to solve (\ref{eq:obj_ori_join_sup}) under the constraint (\ref{eq:binary_join_sup}), similar to SAH (Section~\ref{sec:SAH}), we propose to optimize it with alternating optimizing w.r.t. hashing parameters ($\W,\cc$) and  aggregated representation $\Phi$. 

\textbf{$\Phi$-step:} When fixing $\W_1,\cc_1,\W_2,\cc_2$ and solving for $\Phi$, we can solve over each sample %$\varphi_i$
 independently. Specifically, for each sample $i=1,...,m$, we solve the following relaxed problem by skipping the binary constraint
\begin{eqnarray}
\min_{\varphi_i} \frac{1}{2} \norm{\y_i-\left(\W_2(\W_1\varphi_i+\cc_1)+\cc_2\right)}^2 \nonumber \\ 
{}&&\hspace{-17em}+\frac{\gamma}{2}\left(\norm{\V_i^T\varphi_i-\1}^2+\mu\norm{\varphi_i}^2 \right) \label{eq:obj_pool_sup}
\end{eqnarray}
By solving (\ref{eq:obj_pool_sup}), we find $\varphi_i$ which not only encourages the aggregating property but also encourages a good  classification for a linear classifier, i.e., it minimizes the $l_2$ loss w.r.t. the label vector. (\ref{eq:obj_pool_sup}) is actually a $l_2$ regularized least squares problem, hence we achieve the analytic solution as follows

\vspace{-0.3cm}\footnotesize
\begin{eqnarray}
\varphi_i &=& \left((\W_2\W_1)^T(\W_2\W_1)+\gamma\V_i\V_i^T+\gamma\mu\I\right)^{-1} \nonumber \\ 
{}&&\hspace{0em}\times \left(\gamma\V_i\1+(\W_2\W_1)^T\left(\y_i-(\W_2\cc_1+\cc_2)\right)\right) \label{eq:sol_pool_sup}
\end{eqnarray}
\normalsize
The asymptotic complexity for computing (\ref{eq:sol_pool_sup}) is 
$\mathcal{O} (\max(D^3,D^2n_i))$ 
which is similar to the asymptotic complexity for computing (\ref{eq:gmp-solution}).

\textbf{$(\W,\cc)$-step:} When fixing $\Phi$ and solving for $(\W_1,\cc_1,\W_2,\cc_2)$, (\ref{eq:obj_ori_join_sup}) under the constraint (\ref{eq:binary_join_sup}) is equivalent to the following optimization

\vspace{-0.2cm}\footnotesize
\begin{eqnarray}
\min_{\{\W_i,\cc_i\}_{i=1}^{2}}\frac{1}{2} \norm{\Y-\left(\W_2(\W_1\Phi+\cc_1\1^T)+\cc_2\1^T\right)}^2 \nonumber \\ 
{}&&\hspace{-20em}+\frac{\beta}{2}\left(\norm{\W_1}^2+\norm{\W_2}^2\right) \label{eq:obj_hash_sup}
\end{eqnarray}
\begin{equation}
\textrm{s.t. } \W_1\Phi+\cc_1\1^T \in \{-1,1\}^{L\times m} \label{eq:binary-hash_sup}
\end{equation} 

\normalsize 
By solving (\ref{eq:obj_hash_sup}) under the constraint (\ref{eq:binary-hash_sup}), we find hash parameters which not only ensure the binary outputs of the encoded features but also ensure that the learned binary codes give a good classification, i.e., they minimize the $l_2$ loss  w.r.t. the label.
(\ref{eq:obj_hash_sup}) and (\ref{eq:binary-hash_sup}) have same forms as (\ref{eq:obj_ori}) and (\ref{eq:binary0}) except some changes in variables.  Specifically, the first and the second $\X$ in the first term of (\ref{eq:obj_ori}) are replaced by $\Y$ and $\Phi$, respectively; the variable $\X$ in (\ref{eq:binary0}) is replaced by $\Phi$. 
In spite of changing of some variables between these two formulations, the proposed Relaxed Binary Autoencoder (Section \ref{sec:RBA}) can be used for solving (\ref{eq:obj_hash_sup}) under the constraint (\ref{eq:binary-hash_sup}). In particular, we reuse the Algorithm \ref{alg1} for solving $(\W_1,\cc_1,\W_2,\cc_2)$ in which $\Y$ and $\Phi$ are used as the inputs for the training. 
Note that, by changing the parameters, when using RBA (Algorithm \ref{alg1}), the \textbf{$(\W,\cc)$-step} (Section \ref{subsec:RBA_opt}) is changed as follows

%\vspace{-0.2cm}\small
\begin{equation}
\W_1 = \lambda \left(\B-\cc_1\1^T\right)\Phi^T \left(\lambda\Phi\Phi^T+\beta\I\right)^{-1}
\label{eq:W1_sup}
\end{equation}
\begin{equation}
\W_2 = \left(\Y-\cc_2\1^T\right)\B^T \left(\B\B^T+\beta\I\right)^{-1}
\label{eq:W2_sup}
\end{equation}
%\normalsize 

%\vspace{-0.2cm}
\begin{equation}
\cc_1 = \frac{1}{m}\left(\B-\W_1\Phi\right)\1 
\label{eq:c1_sup}
\end{equation}
\begin{equation}
\cc_2 = \frac{1}{m}\left(\Y-\W_2\B\right)\1
\label{eq:c2_sup}
\end{equation}

Furthermore, at \textbf{$\B$-step} (Section \ref{subsec:RBA_opt}), $\widetilde{\X}$ and $\H$ are computed as follows
\begin{equation}
\widetilde{\X}=\Y-\cc_2\1^T
\label{eq:B_sup}
\end{equation}
\begin{equation}
\H = \W_1 \Phi + \cc_1 
\label{eq:H_sup}
\end{equation}

The learning algorithm SASH is similar to the one of SAH and is presented in the Algorithm \ref{alg3}. In the Algorithm~\ref{alg3}, $\Phi^{(t)}$, $\W_1^{(t)}, \cc_1^{(t)}, \W_2^{(t)}, \cc_2^{(t)}$ are values of the corresponding parameters at $t^{th}$ iteration. 

\begin{algorithm}[!t]
	\footnotesize
	\caption{Simultaneous feature Aggregating and Supervised Hashing (SASH)}
	\begin{algorithmic}[1] 
		\Require 
			\Statex $\Vs=\{\V_i\}_{i=1}^{m}$: training data; $\Y=\{\y_i\}_{i=1}^{m}$: data label; $L$: code length; $T, T_1$: maximum iteration numbers for SASH and RBA (Algorithm \ref{alg1}), respectively; parameters $\lambda, \beta, \gamma, \mu$. 
		\Ensure 
			\Statex 
			Hashing parameters $\W_1, \cc_1, \W_2, \cc_2$ and aggregated representations before and after learning, i.e., $\Phi^{(0)}$, $\Phi^{(T)}$. 
			\Statex 
			\State Initialize $\Phi^{(0)}=\{\varphi_i\}_{i=1}^{m}$ with Generalized Max Pooling (\ref{eq:gmp-solution})
			\For{$t = 1 \to T$}
				\State Fix $\Phi^{(t-1)}$, solve $(\W_1^{(t)}, \cc_1^{(t)}, \W_2^{(t)}, \cc_2^{(t)})$ using Algorithm \ref{alg1} (which uses $\Phi^{(t-1)}$ and $\Y$ as inputs for training, and the \textbf{$(\W,\cc)$-step} and \textbf{$\B$-step} are changed according to (\ref{eq:W1_sup})-(\ref{eq:c2_sup}), and (\ref{eq:B_sup})-(\ref{eq:H_sup}), respectively.)
				\State Fix $(\W_1^{(t)}, \cc_1^{(t)}, \W_2^{(t)}, \cc_2^{(t)})$, solve $\Phi^{(t)}$ using \textbf{$\Phi$-step}.
			\EndFor
			\State Return 
			$\W_1^{(T)}, \W_2^{(T)}, \cc_1^{(T)}, \cc_2^{(T)}, \Phi^{(0)}, \Phi^{(T)}$
    \end{algorithmic}
    \label{alg3}
\end{algorithm}

\subsection{Mapping from original aggregated features to learned aggregated features and binary codes for new image}
%After learning $\W_1, \cc_1, \W_2, \cc_2$, 
\subsubsection{Mapping from original aggregated features to learned aggregated features}
Given set of local features of a test image and learned parameters $(\W_1,\cc_1,\W_2,\cc_2)$, in SASH we can not compute the aggregated representation of the test image using (\ref{eq:sol_pool_sup}) because we do not have the label $\y$ for the test image. To overcome this, we propose a simple yet powerful solution as follows. 
The Algorithm~\ref{alg3} returns the aggregated representations before and after learning of training images, i.e.,  $\Phi^{(0)}$ and $\Phi^{(T)}$. For simple notations, we note them as $\Phi_0$ and $\Phi$.  We compute (only one time in offline) a linear mapping $\P$ which maps $\Phi_0$ to $\Phi$ by solving the following ridge regression

\begin{equation}
\min_{\P} \left( \frac{1}{2}\norm{\Phi - \P\Phi_0}^2  + \frac{\alpha}{2} \norm{\P}^2 \right)
\end{equation}
where the regularization term is to prevent overfitting and to obtain a stable solution.  
We have the analytic solution for $\P$ as follows

\begin{equation}
\P = \Phi \Phi_0^T \left( \Phi_0 \Phi_0^T + \alpha \I \right)^{-1}
\label{eq:mapping}
\end{equation}

\red{It is worth noting that when learning $\mathbf{P}$, the features ${\Phi}$ have been already available, i.e., ${\Phi}$ has been learned by using the label information (on training images). Hence, we can expect that, the aggregated representations $\varphi$ of images from the same class will be close together and they are  also  apart
from the aggregated representations of other classes. In the other words, the aggregated representations of samples from the same class will form a compact cluster. 
That means that $\mathbf{P}$ will map representations $\varphi_0$ of images from the same class to the same cluster, i.e., the distance (between the mapped features) of samples from the same class is expected to be small. This is analogous to the distance metric learning \cite{DBLP:conf/nips/XingNJR02,DBLP:conf/eccv/WenZL016}.  Hence the learning of $P$ can be seen as a simplified distance metric learning which is a well-known approach to the unseen class retrieval   \cite{DBLP:conf/nips/XingNJR02,DBLP:conf/eccv/WenZL016}. 
} 

The mapping $\P$ will be used in the process of computing binary codes for new images as the following. 
\subsubsection{Binary codes for new image}
Given set of local features of a new image, we first compute its GMP representation $\varphi_0$ using (\ref{eq:gmp-solution}). We then compute its  learned aggregated representation $\varphi$ by 
\begin{equation}
\varphi = \P\varphi_0
\label{eq:varphi}
\end{equation}
where $\P$ is defined by (\ref{eq:mapping}). After that, we pass $\varphi$ to the learned encoder %i.e., $h = \W_1\varphi+\cc_1$, 
to compute the binary code. \\
\textbf{Asymptotic complexity: }
When computing the binary codes for a new image, it involves three steps, i.e., \textit{(i)} computing $\varphi_0$ using (\ref{eq:gmp-solution}); \textit{(ii)} computing $\varphi$ using (\ref{eq:varphi}); \textit{(iii)} computing binary codes by the encoder $sgn(\W_1\varphi+\cc_1)$. The asymptotic complexities for these three steps are $\mathcal{O} (\max(D^3,D^2n_i))$, $\mathcal{O}(D^2)$, and $\mathcal{O}(LD)$,  respectively. We can see that the most expensive step is to compute $\varphi_0$.  
Hence, the asymptotic complexity when computes $\varphi$ is similar to the one of GMP (\ref{eq:gmp-solution}). 

\section{Evaluation of Simultaneous Feature Aggregating and Supervised Hashing (SASH)}
\label{sec:evaSASH}

In this section, we evaluate and compare the proposed SASH with state-of-the-art supervised hashing methods including Supervised Discrete Hashing (SDH)~\cite{Shen_2015_CVPR}, ITQ-CCA~\cite{DBLP:conf/cvpr/GongL11}, Kernel-based Supervised Hashing (KSH)~\cite{CVPR12:Hashing}, Binary Reconstructive Embedding (BRE)~\cite{Kulis_learningto}. For all compared methods, we use the implementations and the suggested parameters provided by the authors. We also compare the proposed SASH to recent end-to-end deep supervised hashing methods, i.e., 
Deep Quantization Network (DQN)~\cite{DBLP:conf/aaai/CaoL0ZW16}, Deep Hashing Network (DHN)~\cite{DBLP:conf/aaai/ZhuL0C16}, Deep Supervised Discrete Hashing (DSDH)~\cite{DSDH}. 
For the proposed SASH, the values of $\lambda$ (in~Eq. (\ref{eq:W1_sup})), $\beta$ (in~Eq. (\ref{eq:W1_sup})), and $\alpha$ (in~Eq. (\ref{eq:mapping})) %in (\ref{eq:obj_ori_join})
  are respectively set by cross validation as $10^{-4}$, $10^{-3}$, and $5\times10^{-1}$ for all experiments. We cross-validate $\gamma$ and $\mu$ (in the objective function~(\ref{eq:obj_ori_join_sup}))  in the ranges of $[10^{-1}, 10]$%$[10^{-2}, 1]$ 
  and $[10^1, 10^{5}]$, respectively, with the multiplicative step-size of 10.

\subsection{Dataset}
Follow the state of the art~\cite{Shen_2015_CVPR,CVPR12:Hashing,DBLP:conf/cvpr/LaiPLY15}, we evaluate the proposed method on the standard supervised hashing benchmarks CIFAR10~\cite{Krizhevsky09},  MNIST~\cite{mnistlecun} and NUS-WIDE~\cite{nus-wide-civr09}.  
The descriptions of CIFAR10, MNIST datasets have been presented in the Section~\ref{subsub_eva_rba}. The description of NUS-WIDE dataset is provided in the following. 
In order to configure the training and testing splits, we follow both traditional configuration and a recent proposed configuration~\cite{DBLP:conf/icassp/SablayrollesDUJ17}. \\
\subsubsection{Traditional configuration}
We follow the traditional configuration in state-of-the-art supervised hashing methods~\cite{CVPR12:Hashing,Shen_2015_CVPR}.

\textbf{CIFAR10} for the CIFAR10 dataset, we randomly sample 100 images per class to form 1,000 query images. The remaining 59K images are used as database images. Furthermore, 500 images per class are sampled from the database to form 5K training images. \\

\textbf{MNIST} for the MNIST dataset, we randomly sample 100 images per class to form 1,000 query image. The remaining images are used as the training images and database images. Note that as KSH and BRE require a full similarity matrix when training, it is difficult for these methods to handle large training data. Follow SDH~\cite{Shen_2015_CVPR}, we sample 5,000 images for training these methods. \\

\textbf{NUS-WIDE } \cite{nus-wide-civr09} dataset contains about 270,000 images collected from Flickr. NUS-WIDE is associated with
81 ground truth labels, with each image containing multiple semantic labels. We define the groundtruths of a query as the images sharing at least one label with the query image. As in~\cite{CVPR12:Hashing,Shen_2015_CVPR}, we select the 21 most frequent labels. 
For each label, we randomly sample 100 images for the query set. The remaining images are used as database images. Furthermore, 500 images per class are randomly sampled from the database to form the training set.

\subsubsection{``Retrieval of unseen classes'' configuration}
Recently, in~\cite{DBLP:conf/icassp/SablayrollesDUJ17}, the authors show that in the  traditional configuration, by using same class labels for both training and testing phases, that configuration is more related to the classification task, rather than the retrieval task. Hence, to evalute the proposed method for the retrieval context, which is the focus of this paper, we also follow the ``retrieval of unseen classes'' configuration that is proposed in~\cite{DBLP:conf/icassp/SablayrollesDUJ17}. 
For CIFAR10 and MNIST, we start from their original splits (i.e., the number of training and testing images are 50K and 10K for CIFAR10, 60K and 10K for MNIST) but we use separate classes at training and testing stages. Specifically, $70\%$ of the class labels, which are randomly sampled, are used when learning the hashing function, and the $30\%$ remaining class labels (unseen classes) are used to evaluate the hashing scheme. We call train70/test70 the train/test images of the $70\%$ classes and train30/test30 the remaining ones. The train70 is used to train the hash function. The test30 is used as queries. The train30 is used as database for the retrieval.  In summary, the train70 - train30 - test30 is equivalent to the learn - database - query. The test70 is not used at all.

For NUS-WIDE dataset, we start from the data split of the traditional configuration, i.e. the testing set consists of 100 images per class, the remaining images form database. We  randomly split the 21 most frequent labels into 2 groups of $70\%$ (i.e. 15) classes for training and $30\%$ (i.e. 6) classes for unseen retrieval testing. As a result, we have database70/database30 and test70/test30 sets.  
Note that since each image may have multiple labels, we post-process the test30 and database30 by removing images that contain labels appeared in the set of $70\%$ seen classes. This ensures that there is no overlap in the class labels between training images and testing images. We then sample 500 images for each class from database70 to form train70 set, which is used to train the hash function. The test30 and database30 are respectively used as query set and database for testing.

\begin{table}[!t]
   \centering
   \footnotesize
   %\small
   \caption{mAP (\%) comparison between SASH and state-of-the-art supervised hashing methods on CIFAR10 dataset under traditional configuration.}
    \begin{tabular}{|l|c|c|c|c|c|} 
    \hline
  L(bits)    &8  &16  &24  &32  &48  \\ \hline
CCA-ITQ\cite{DBLP:conf/cvpr/GongL11}&35.44	&38.85	&41.21	&43.77 &45.30 \\ \hline
KSH\cite{CVPR12:Hashing}  &32.80	&37.55	&39.13	&40.57	&41.81\\ \hline
BRE\cite{Kulis_learningto}  &19.15	&22.37	&24.11	&25.59	&26.50\\ \hline
SDH\cite{Shen_2015_CVPR} &40.35	&43.83	&45.92	&47.56	&48.69\\ \hline
SASH &53.57	&57.45	&61.33	&62.82	&63.65 \\ \hline
    \end{tabular}
    \label{tab:SASH_cifar}
\end{table}

\begin{table}[!t]
   \centering
   \footnotesize
   %\small
   \caption{mAP (\%) comparison between SASH and state-of-the-art supervised hashing methods on MNIST dataset under traditional configuration.}
    \begin{tabular}{|l|c|c|c|c|c|} 
    \hline
  L(bits)    &8  &16  &24  &32  &48  \\ \hline
CCA-ITQ\cite{DBLP:conf/cvpr/GongL11} &58.18	&64.24	&68.07	&69.42	&70.92 \\ \hline
KSH\cite{CVPR12:Hashing}  &65.40	&73.81	&76.09	&78.44	&79.52\\ \hline
BRE\cite{Kulis_learningto}  &27.17	&32.27	&38.38	&40.24	&42.37\\ \hline
SDH\cite{Shen_2015_CVPR}&69.23	&76.31	&78.12	&80.91	&81.63\\ \hline
SASH &75.48	&79.34	&81.02	&83.63	&84.51\\ \hline
    \end{tabular}
    \label{tab:SASH_mnist}
\end{table}

\begin{table}[!t]
   \centering
   \footnotesize
   %\small
   \caption{mAP (\%) comparison between SASH and state-of-the-art supervised hashing methods on NUS-WIDE dataset under traditional configuration.}
    \begin{tabular}{|l|c|c|c|c|c|} 
    \hline
  L(bits)    &8  &16  &24  &32  &48  \\ \hline
CCA-ITQ\cite{DBLP:conf/cvpr/GongL11} &56.59	&59.98	&61.45 &61.75  &62.45\\\hline
KSH\cite{CVPR12:Hashing} &62.97	& 66.51	& 67.42 & 67.59 & 67.94\\ \hline
BRE\cite{Kulis_learningto}&34.87 &36.34	&37.19	&38.14 &39.94 \\ \hline
SDH\cite{Shen_2015_CVPR} &63.07	&66.89  &67.36 &67.79	&68.10\\ \hline
SASH & 64.01	& 67.10	& 67.63	& 68.01 & 68.75\\ \hline
    \end{tabular}
    \label{tab:SASH_NUS-WIDE}
\end{table}

\subsection{Evaluation protocol}
As standardly done in the literature, the retrieval accuracy is reported in term of mean Average Precision (mAP). For all datasets, the labels of images are used as the groundtruth. 

Similar to SAH in Section~\ref{subsub:cnnfea}, we use the $5^{th}$ convolutional features of the pre-trained VGG network \cite{Simonyan14c} as the inputs for the proposed SASH. In order to make a fair comparison between SASH and other hashing methods, i.e., KSH, BRE, CCA-ITQ, SDH, we aggregate the convolutional features with GMP \cite{gmp} and use the resulted vectors as the inputs for compared hashing methods. 

\begin{table}[!t]
\centering
\footnotesize
%\small
\caption{mAP (\%) comparison between SASH and state-of-the-art supervised hashing methods on CIFAR10  dataset under ``retrieval of unseen classes'' configuration.}
\begin{tabular}{|l|c|c|c|c|c|} 
\hline
L(bits)    &8  &16  &24  &32  &48  \\ \hline
CCA-ITQ\cite{DBLP:conf/cvpr/GongL11} & 65.48 & 69.94 & 70.72 & 71.33 & 71.83\\ \hline
KSH\cite{CVPR12:Hashing}& 62.41 & 64.11 & 62.13 & 64.40 & 62.61 \\ \hline
BRE\cite{Kulis_learningto}& 52.55 & 51.20 & 53.10 & 55.73 & 54.34\\ \hline
SDH\cite{Shen_2015_CVPR}& 47.43 & 58.41 & 61.83 & 60.08 & 62.60\\ \hline
SASH & 70.08 & 74.46 & 77.22 & 78.33 & 79.86  \\ \hline
\end{tabular}
\label{tab:SASH_cifar_unseen}
\end{table}

\begin{table}[!t]
\centering
\footnotesize
%\small
\caption{mAP (\%) comparison between SASH and state-of-the-art supervised hashing methods on MNIST  dataset under ``retrieval of unseen classes'' configuration.}
\begin{tabular}{|l|c|c|c|c|c|} 
\hline
L(bits)    &8  &16  &24  &32  &48  \\ \hline
CCA-ITQ\cite{DBLP:conf/cvpr/GongL11} & 52.40 & 56.36 & 58.71 & 59.82 & 60.25 \\ \hline
KSH\cite{CVPR12:Hashing}& 50.27 & 55.64 & 57.21 & 58.31 & 58.85\\ \hline
BRE\cite{Kulis_learningto}& 50.03 & 53.43 & 55.52 & 56.91 & 57.11\\ \hline
SDH\cite{Shen_2015_CVPR}& 53.64 & 55.80 & 57.31 & 58.14 & 59.69\\ \hline
SASH & 59.10 & 62.20 &63.31 & 63.94 & 64.21 \\ \hline
\end{tabular}
\label{tab:SASH_mnist_unseen}
\end{table}

\begin{table}[!t]
\centering
\footnotesize
%\small
\caption{mAP (\%) comparison between SASH and state-of-the-art supervised hashing methods on NUS-WIDE dataset under ``retrieval of unseen classes'' configuration. 
}
\begin{tabular}{|l|c|c|c|c|c|} 
\hline
L(bits)    &8  &16  &24  &32  &48  \\ \hline
CCA-ITQ\cite{DBLP:conf/cvpr/GongL11} & 37.81 & 42.93 & 43.92 & 44.31 & 44.85\\ \hline
KSH\cite{CVPR12:Hashing} & 36.18 & 38.84 & 39.88 & 40.08 & 40.98\\ \hline
BRE\cite{Kulis_learningto}  & 34.10 & 33.95 & 35.22 & 36.14 & 35.79\\ \hline
SDH\cite{Shen_2015_CVPR} & 34.92 & 43.45 & 43.35 & 43.89 & 44.55\\ \hline
SASH & 42.62 & 44.26 & 45.05 & 45.43 & 46.71 \\ \hline
\end{tabular}
\label{tab:SASH_NUS-WIDE_unseen}
\end{table}

\subsection{Retrieval results}

\begin{table*}[!t]
\centering
%\footnotesize
\caption{mAP (\%) comparison between SASH and end-to-end deep learning-based supervised hashing methods on CIFAR10, MNIST, and NUS-WIDE under ``retrieval of unseen classes'' configuration. 
}
\begin{tabular}{|l|c|c|c|c|c|c|c|c|c|} 
\hline
\multirow{2}{*}{} & \multicolumn{3}{c|}{CIFAR10} & \multicolumn{3}{c|}{MNIST} & \multicolumn{3}{c|}{NUS-WIDE} \\
%\cline{1-7}
\hline	L(bits)   &24  &32  &48   &24  &32    &48  &24  &32    &48  \\ \hline 
DQN\cite{DBLP:conf/aaai/CaoL0ZW16}   
& 75.15	& 76.05	& 76.78 
& 59.68	& 61.32	& 63.76 
& 43.37 & 43.78 & 44.24 \\\hline
DHN\cite{DBLP:conf/aaai/ZhuL0C16}    
& 74.53 & 75.95 & 76.59 
& 60.10 & 62.57	& 63.43 
& 44.65 & 45.03 & 44.98 \\\hline

DSDH \cite{DSDH} & 74.37 & 75.41 & 76.23 & 62.78 & 63.18 & 64.05 & 39.05 & 40.02 & 40.58 \\\hline
SASH	
& 77.22	& 78.33	& 79.86 
& 63.31 & 63.94 & 64.21
& 45.05 & 45.53 & 46.71 \\\hline
\end{tabular}
\label{tab:SASH_deep_unseen}
\end{table*}
\subsubsection{Results on the traditional configuration}
Tables~\ref{tab:SASH_cifar} and \ref{tab:SASH_mnist} present comparative results between the proposed SASH and compared supervised hashing methods on the CIFAR10 and MNIST datasets under the traditional configuration. The results show that SASH significantly outperform the compared methods on both datasets. The most competitive method is SDH~\cite{Shen_2015_CVPR}. The improvement of SASH over SDH is more clear on the CIFAR10 dataset, i.e., from 13\%-15\% at different code lengths. 
Table~\ref{tab:SASH_NUS-WIDE} presents comparative results on the NUS-WIDE dataset. The results show that the proposed SASH, SDH~\cite{Shen_2015_CVPR}, and KSH~\cite{CVPR12:Hashing} achieve competitive results. Specifically, SASH achieves slightly higher performances than SDH and KSH, and these three methods significantly outperform other methods, e.g., CCA-ITQ, BRE. 

\subsubsection{Results on the ``retrieval of unseen classes'' configuration}
Tables~\ref{tab:SASH_cifar_unseen} and \ref{tab:SASH_mnist_unseen} present comparative results between the proposed SASH and compared supervised hashing methods on the CIFAR10 and MNIST datasets under the ``retrieval of unseen classes'' configuration. Under this configuration, the proposed SASH  outperforms compared methods a fair margin. The most competitive method is CCA-ITQ~\cite{DBLP:conf/cvpr/GongL11}. On the CIFAR10 dataset, SASH outperforms CCA-ITQ around 4.5\% to 8\% at different code lengths. 
On the MNIST dataset, the improvements of SASH over CCA-ITQ vary around  4\% to 6.5\% at different code lengths. 
Table~\ref{tab:SASH_NUS-WIDE_unseen} presents comparative results on the NUS-WIDE dataset. The proposed SASH also outperforms other methods. However, the improvements are lower than those on CIFAR10 and MNIST datasets, i.e., the improvements of SASH over CCA-ITQ vary around 1\% to 5\% at different code lengths. The higher improvements are observed at low code lengths, e.g., $L=8$. \\

\textbf{Comparison to end-to-end deep supervised hashing methods:} 
Most traditional end-to-end deep supervised hashing methods~\cite{DBLP:conf/cvpr/LaiPLY15,DBLP:conf/aaai/CaoL0ZW16,DBLP:conf/aaai/ZhuL0C16,DSDH} consist of the fine-tuning a deep Convolutional Neural Network which is trained for the classification task. As shown in~\cite{DBLP:conf/icassp/SablayrollesDUJ17}, under the traditional configuration in which the training and testing use the same class labels, one can directly encode the output of the classification layer (e.g. a softmax layer) using only $\lceil log_2 C \rceil$ bits, where $C$ is the number of classes. This simple strategy actually outperforms state-of-the-art supervised hashing methods. However, that configuration is suitable for classification problem, not retrieval problem. Hence, here we evaluate and compare the proposed SASH with end-to-end deep hashing methods using ``retrieval of unseen classes'' setting~\cite{DBLP:conf/icassp/SablayrollesDUJ17}. 

Table~\ref{tab:SASH_deep_unseen} presents comparative results between the proposed SASH and recent end-to-end supervised deep hashing methods DQN~\cite{DBLP:conf/aaai/CaoL0ZW16}, DHN~\cite{DBLP:conf/aaai/ZhuL0C16}, DSDH~\cite{DSDH}. It is worth noting that in the original corresponding papers, the authors did not report the performance of these methods under unseen class setting. Hence we use the released implementations and the suggested parameters provided by the authors to conduct experiments. Follow~\cite{DBLP:conf/aaai/CaoL0ZW16,DBLP:conf/aaai/ZhuL0C16}, we report the results with $L=24, 32$ and $48$ bits. 
The experimental results show that on the simple MNIST dataset, SASH achieves competitive results to DQN~\cite{DBLP:conf/aaai/CaoL0ZW16} and DHN~\cite{DBLP:conf/aaai/ZhuL0C16} at higher code lengths, i.e., $L=32$ and $48$, while it outperforms these two methods at lower code length, i.e., $L=24$. SASH and DSDH~\cite{DSDH} achieve competitive results at all code lengths on the MNIST dataset. 
On the CIFAR10 dataset, SASH clearly outperforms DQN, DHN, DSDH fair margins at all compared code lengths, i.e., SASH outperforms the second best  DQN around 2\% to 3\% at different code lengths. 
Regarding the NUS-WIDE dataset, the proposed SASH  achieves favorable performances over DQN and DHN across all code lengths, while it significantly outperforms DSDH at all compared code lengths.

\section{Conclusion}
\label{sec:concl}

In this paper, we first introduce the Relaxed Binary Autoencoder (RBA) hashing method in which we obtain analytic solutions for encoder and decoder during the alternating learning. This leads to the efficient training of RBA. After that, we propose a novel unsupervised hashing approach, i.e., SAH, 
in which the feature aggregating and hashing are designed simultaneously and optimized jointly. The binary codes are learned such that they not only encourage the aggregating property but also ensure a good reconstruction of the inputs. We further propose a supervised version of SAH, namely, SASH, by leveraging the label information when learning binary codes. The binary codes are learned such that they not only encourage the aggregating property but also optimize for a linear classifier. 
Extensive experiments on benchmark datasets with different image features and different configurations demonstrate that the proposed methods outperforms the state-of-the-art unsupervised and supervised hashing methods.

\section*{Acknowledgement}

This work was supported by both ST Electronics and the National Research Foundation(NRF), Prime Minister's Office, Singapore under Corporate Laboratory @ University Scheme (Programme Title: STEE Infosec - SUTD Corporate Laboratory).

%\appendices
%\input{append}

%% use section* for acknowledgment
%\ifCLASSOPTIONcompsoc
%  % The Computer Society usually uses the plural form
%\else
%  % regular IEEE prefers the singular form
%\fi

\ifCLASSOPTIONcaptionsoff
  \newpage
\fi

{%\small
\bibliographystyle{IEEEtran}
\bibliography{hash}
}

\vspace{-1cm}
\begin{IEEEbiography}[{\includegraphics[width=0.9in,height=1.1in,clip,keepaspectratio]{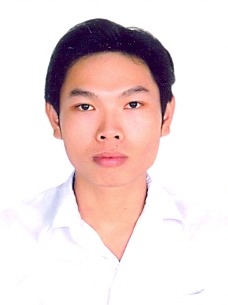}}]
%\begin{IEEEbiography}[]
{Thanh-Toan Do}
is  currently  a Lecturer at the Department of Computer Science, the University of Liverpool (UoL), United Kingdom. He obtained  Ph.D.  in  Computer  Science  from  INRIA, Rennes, France in 2012. Before joining UoL, he was a  Research  Fellow  at  the  Singapore  University  of Technology  and  Design,  Singapore  (2013  -  2016) and  the  University  of  Adelaide,  Australia  (2016  - 2018).  His  research  interests  include  Computer  Vision and Machine Learning.
\end{IEEEbiography}

\vspace{-1cm}
\begin{IEEEbiography}[{\includegraphics[width=1in,height=1.2in,clip,keepaspectratio]{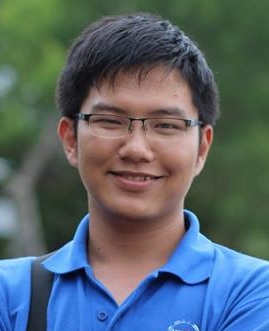}}]
%\begin{IEEEbiography}[]
{Khoa Le}
 received the BSc for an honours degree from the University of Science, Vietnam National University, in 2015. Since 2016, he has been a research assistant at Singapore University of Technology and Design (SUTD). His current research interests are deep learning and image retrieval.
\end{IEEEbiography}

\vspace{-1.5cm}
\begin{IEEEbiography}[{\includegraphics[width=1in,height=1.25in,clip,keepaspectratio]{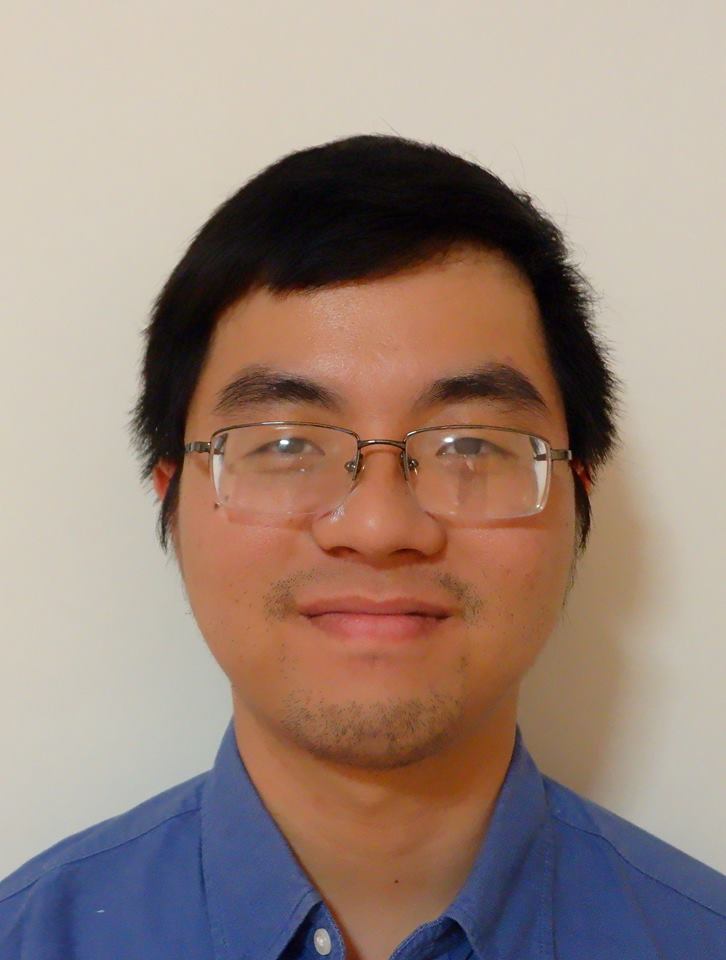}}]
%\begin{IEEEbiography}[]
{Tuan Hoang}
is currently a Ph.D. student at Singapore University of Technology and Design (SUTD) Jan 2016. Before joining SUTD, he achieved the bachelor degree in Electrical Engineering at Portland State University, in 2014. His research interests are content-based image retrieval and image hashing.
\end{IEEEbiography}

\vspace{-1.5cm}
\begin{IEEEbiography}[{\includegraphics[width=1in,height=1.25in,clip,keepaspectratio]{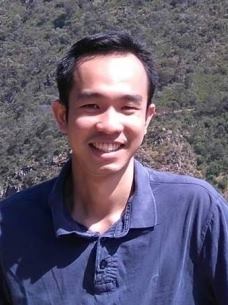}}]
%\begin{IEEEbiography}[]
{Huu Le}
 received the B.S. degree in Electrical and Computer Engineering from Portland State University, Oregon, USA, in 2011 and the Ph.D. degree in Computer Science from the University of Adelaide, Australia, in 2018. He is currently a postdoctoral researcher at Chalmers University of Technology, Sweden. His research interests include robust estimation, non-rigid registration, computational geometry, large-scale image retrieval and optimization methods applied to the fields of computer vision.
\end{IEEEbiography}

\vspace{-1.5cm}
\begin{IEEEbiography}[{\includegraphics[width=1in,height=1.25in,clip,keepaspectratio]{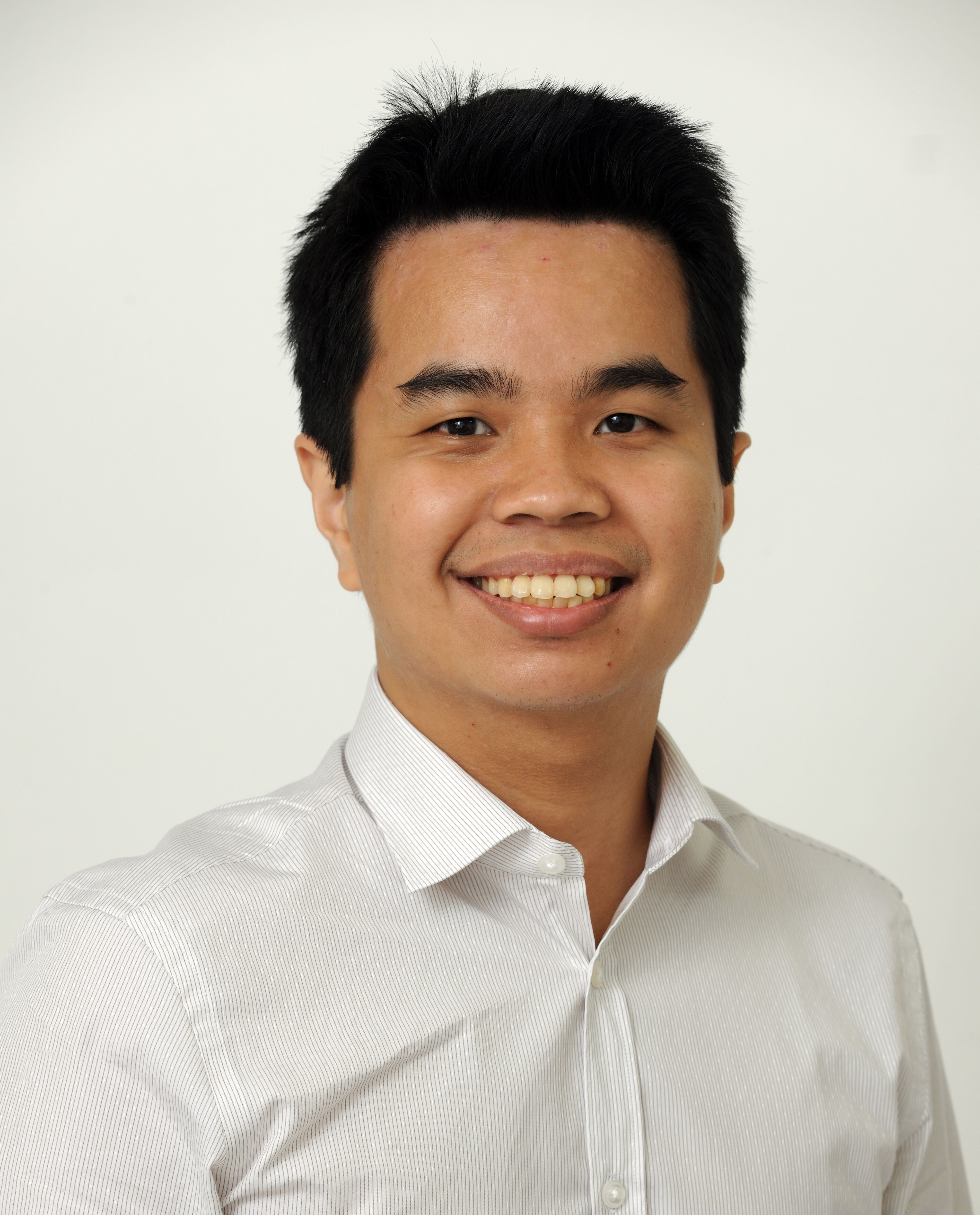}}]
%\begin{IEEEbiography}[]
{Tam V. Nguyen}
  is an Assistant Professor at Department of Computer Science, University of Dayton. He received PhD degree in National University of Singapore (NUS) in 2013. His research topics include computer vision, applied deep learning, multimedia content analysis, and mixed reality. He has authored and co-authored 50+ research papers with 900+ citations according to Google Scholar.  His works were published at IJCV, IEEE T-IP, IEEE T-MM, IEEE T-CSVT, Neurocomputing, ECCV, IJCAI, AAAI, and ACM Multimedia. He is an IEEE Senior Member.
\end{IEEEbiography}

\vspace{-1.5cm}
\begin{IEEEbiography}[{\includegraphics[width=1in,height=1.25in,clip,keepaspectratio]{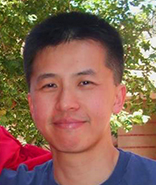}}]
%\begin{IEEEbiography}[]
{Ngai-Man Cheung}
received the Ph.D. degree in electrical engineering from the University  of Southern California, Los Angeles, CA, in 2008. He is currently an Associate Professor with the Singapore University of Technology and Design (SUTD). From 2009 - 2011, he was a postdoctoral researcher with the Image, Video and Multimedia Systems group at Stanford University, Stanford, CA.  He has also held research positions with Texas Instruments Research Center Japan, Nokia Research Center, IBM T. J. Watson Research Center, HP Labs Japan, Hong Kong University of Science and Technology (HKUST), and Mitsubishi Electric Research Labs (MERL). His work has resulted in 10 U.S. patents granted with several pending. His research interests include signal, image, and video processing, and computer vision.
\end{IEEEbiography}

\end{document}